\documentclass[Afour,sageh,times]{sagej}

\usepackage{graphicx} 
\usepackage{moreverb,url}
\usepackage[colorlinks,linkcolor=darkgray,
	    urlcolor=darkgray,citecolor=darkgray,
	    anchorcolor=darkgray,bookmarksopen,
	    bookmarksnumbered]{hyperref}
\usepackage[TS1,T1]{fontenc}
\usepackage{amsmath,amssymb}
\usepackage{longtable}
\usepackage[utf8x]{inputenc}
\usepackage[table]{xcolor}
\usepackage{array}
\usepackage{float}
\usepackage{tikz}
\usetikzlibrary{positioning}
\usepackage{tabularx}
\usepackage[normalem]{ulem}
\usepackage{xtab,booktabs}
\usepackage{csquotes}

\newcommand\BibTeX{{\rmfamily B\kern-.05em \textsc{i\kern-.025em b}\kern-.08em
T\kern-.1667em\lower.7ex\hbox{E}\kern-.125emX}}

\newcommand{\etal}{\mbox{\emph{et al.\ }}}

\newcommand\pzout{\bgroup\markoverwith
{\textcolor{red}{\rule[.5ex]{2pt}{0.4pt}}}\ULon}
\newcommand\shout{\bgroup\markoverwith
{\textcolor{blue}{\rule[.5ex]{2pt}{0.4pt}}}\ULon}
\newcommand\jpout{\bgroup\markoverwith
{\textcolor{orange}{\rule[.5ex]{2pt}{0.4pt}}}\ULon}
\newcommand\erout{\bgroup\markoverwith
{\textcolor{cyan}{\rule[.5ex]{2pt}{0.4pt}}}\ULon}
\newcommand\xzout{\bgroup\markoverwith
{\textcolor{green}{\rule[.5ex]{2pt}{0.4pt}}}\ULon}

\usepackage[normalem]{ulem}

\def\volumeyear{2016}

\hypersetup{
	pdftitle={Action Representations in Robotics: A Taxonomy and Systematic Classification},
	pdfauthor={Philipp Zech, Erwan Renaudo, Simon Haller, Xiang Zhang, Justus Piater}
}

\begin{document}

\runninghead{Zech et al.}

\title{Action representations in robotics: A taxonomy and systematic classification}

\author{Philipp Zech, Erwan Renaudo, Simon Haller, Xiang Zhang and Justus Piater}

\affiliation{Department of Computer Science, University of Innsbruck, Tyrol, Austria}

\corrauth{Philipp Zech, Intelligent and Interactive Systems Group.\\
Department of Computer Science, University of Innsbruck\\
Technikerstrasse 21a \\
Innsbruck, Tyrol\\
Austria}

\email{philipp.zech@uibk.ac.at}

\begin{abstract}
	Understanding and defining the meaning of ``action'' is substantial for robotics research. This
	becomes utterly evident when aiming at equipping autonomous robots with robust
	manipulation skills for action execution. Unfortunately, to this day we still lack
	both a clear understanding of the concept of an action and a set of established
	criteria that ultimately characterize an action. In this survey we thus first review
	existing ideas and theories on the notion and meaning of action. Subsequently we
	discuss the role of action in robotics and attempt to give a seminal definition of
	action in accordance with its use in robotics research. Given this definition we
	then introduce a taxonomy for categorizing action representations in robotics along
	various dimensions. Finally, we provide a systematic literature survey on action
	representations in robotics where we categorize relevant literature along our taxonomy.
	After discussing the current state of the art we conclude with an outlook towards
	promising research directions.
\end{abstract}

\keywords{Action Representations, Robotics}

\maketitle


\section{Introduction}

\emph{In the beginning was the action}\footnote{Und schreibe getrost: im Anfang war die That!}~\cite[p.~81]{Goethe1808}.
Inspired by the Gospel of John, Goethe used this nowadays famous quotation in
the third scene, first act of his famous play ``Faust''. Like Dr.~Faust who
back then struggled with a proper translation for the Greek word ``logos'',
similarly we nowadays struggle with the exact meaning of the word ``action''.
Despite various attempts at formalizing the notion of an action early in this decade,
e.g.,~\cite{Davidson2001} or~\cite{Jeannerod2006},
the controversy on the exact nature of action is still active (see Section~\ref{sec:2}).
Clearly, such a lack of understanding and of an accepted definition hampers research related to understanding
human actions, e.g., in neuroscience or psychology, but also computational descriptions of action, e.g., in the field of robotics research.

\cite{Kruger2007} published a thorough review on
action recognition and mapping in the fields of computer vision, robotics
and artificial intelligence. They however stop short of providing a clear definition
of action itself. Yet, Kr\"uger~\etal already provide a preliminary discussion
of some criteria relevant for characterizing the notion of action. In our work,
we build on these criteria (see Section~\ref{sec:3}).

More recently,~\cite{Weinland2011} published a survey on 
vision-based methods for action representation, segmentation and recognition. Despite 
providing a thorough overview of existing approaches, their survey is limited to categorizing approaches according to their (i) spatial representation, 
(ii) temporal model, (iii) temporal segmentation, and (iv) view-independent representation. 
In contrast, in our work we aim to categorize action representations along many more dimensions 
(see Section~\ref{sec:3}). Further, Weinland~\etal do not provide an underlying definition 
of action as a foundation for their classification. Last but not least, Weinland~\etal do not 
consider the notion of an action's effect which not only since~\cite{Jeannerod2006} 
is considered an integral aspect of an action representation but already dates back at least to~\cite{Bernstein1996}.

The goal of our survey is to define classification criteria that are instrumental for a 
formal treatment of action representations in robotics. We thus aim at capturing the notion 
of action over a sufficiently broad range of analytical viewpoints that have emerged from 
both their theoretical interrogation but also from practical applications. We further 
present a thorough investigation of existing action-related research in
robotics by categorizing relevant publications according to these criteria in a systematic way
(see Section~\ref{sec:4}). As a result of this classification we then provide a comprehensive
and qualitative discussion of existing research to identify both promising and potentially
futile directions as well as open problems and research questions to be addressed in the future
(see Section~\ref{sec:6}). To the best of our knowledge our work is seminal in both introducing
a taxonomy for action representations in robotics and an in-depth discussion of existing research
motivated by a quantitative study.

\paragraph*{Contribution} The core contribution of this article is the introduction of a
comprehensive taxonomy for categorizing action representations in robotics. A systematic 
literature search (see Section~\ref{sec:4}) of the keywords \emph{action} and \emph{representation} 
resulted in 1575 hits, which were systematically reduced to 469
considered papers. Out of those, we identified and categorized 152 major
contributions in the field of robotics. For each publication it was possible to categorize
the employed action representation as applicable. Given the resulting classification we then discuss
the current state of the art of action representation in robotics (see Section~\ref{sec:5}). Finally, on the basis
of this discussion, we identify promising directions for future research (see Section~\ref{sec:6}).

\paragraph*{Intentional Limitations} In this survey, only action representations that have
an application in the field of robotics will be considered. Apart from that, we avoid 
categorizing papers that just build on existing models (see Section~\ref{sec:4}). 
Another limitation we impose on our survey is the 
deliberate exclusion of any papers or articles discussing plain controllers for implementing 
some movement. Though one could consider such a controller an action representation in some 
sense by arguing that it represents an ``action'' by its goal, i.e., a setpoint, we argue 
that controllers do not comprise an action representation simply by missing most of the 
aspects discussed in Section~\ref{sec:3}.


\section{What is an action?}\label{sec:2}

Despite being subtle in its form, the question of \emph{what is an action} has a
long history and probably first was investigated by Aristotle in his study on animal 
movement \emph{De motu animalius}, where he contends that actions are justified as 
of a logical connection between goals and knowledge of effects~\citep{Russell2016,aristotleb},
\begin{quotation}
	But how does it happen that thinking is sometimes accompanied by actions and sometimes 
	not, sometimes by motion, and sometimes not? It looks as if almost the same thing 
	happens in case of reasoning and making inference about unchanging objects. But in 
	that case the end is a speculative proposition \dots whereas here the conclusion is 
	which results from the two premises is an action \dots I need covering; a cloak is 
	covering. I need a cloak. What I need, I have to make; I need a cloak. I have to make 
	a cloak. And the conclusion, the "I have to make a cloak" is an action.
\end{quotation}
Aristotle pursued his studies further in his third 
book of the \emph{Nicomachean Ethics} \citep{aristotle}. In his treatise---though now 
primarily focusing on ethics by attempting to answer the Socratic question of how men
should best live---Aristotle already apprehended the imperative notion of human actions
by attributing them a primary role in shaping a virtuous character. He thence
introduces three categories of actions relevant to virtue, but also whether they
are to be blamed, forgiven or even pitied:
\begin{itemize}
	\item\emph{Voluntary actions} are the righteous actions done by choice, i.e., on
		purpose. They result in increased happiness (\emph{eudaimonia}).
	\item\emph{Involuntary} or \emph{unwilling actions} are neither praised nor blamed as
		in such cases no wrong action is chosen. This strongly builds on ignoring  of what aims
		are good and bad.
	\item\emph{Non-voluntary} or \emph{non-willing actions} are bad actions done by choice,
		i.e., on purpose. They are preferred as all remaining options would be worse.
\end{itemize}
Admittedly, Aristotle did not discuss more specifically what an action is and
also how it may be represented in our minds. Nevertheless, his thoughts are essential
by clearly outlining different types of actions, thus ultimately implying that there
must exist some internal representation which allows choosing among which action to do
given a deliberate purpose. In contrast, if all actions are just hard-coded motor responses
to external stimuli and no higher-level cognitive planning would precede action execution,
such internal representations of actions would be pointless.

\subsection{Action in psychology}

In his article \emph{Action-oriented representation},~\cite{Mandik2005} discusses the nature
of mental representations. Motivated by decade-lasting discussions between proponents
of both underdetermined and determined (or active) perception, Mandik presents arguments
from both conservative embodied cognition (CEC;~or representationalism) and radically
embodied cognition (REC) towards the nature of an internal representation of perception
culminating in what he calls \emph{action-oriented representation} (AOR).

Classically, the school of CEC calls for the need of an internal mental representation.
This theory may be roughly identified as~\cite[p.~287]{Mandik2005}
\begin{quotation}
	[\dots] the view that one has a perceptual experience of an \emph{F} if and only if
	one mentally represents that an \emph{F} is present and the current token mental
	representation of an F is causally triggered by the presence of an \emph{F}.
\end{quotation}
Mandik then argues that the representationalist analysis of perception yields two crucial
components: the \emph{representational} component and the \emph{causal} component. Whereas
the former's job is to account for the similarity between perception on the one hand
and imagery and illusion on the other hand, the latter is required to articulate the idea
that in spite of similarities, there are crucial differences between perceptions and
other representational mental phenomena (e.g., the relevant mental representation of
an \emph{F} must be caused by an \emph{F} to count as percept of an \emph{F}; \citealt{Mandik2005}).

REC on the contrary argues against the explicit need for internal representations by
relying on active perception. This essentially capitalizes on a perception-action cycle
on the sensori-motor level in that actions are directly triggered by stimuli in the environment
without the need for internal representations (\emph{c.f.}~\citealt{Gibson1966,Gibson1979}).
Mandik argues however that active perception can be explained in terms of the representational
theory of perception by acknowledging~\cite[p.~292]{Mandik2005}

\begin{quotation}[\dots] that there are occasions in which outputs instead of inputs figure into
	the specification of the content of a representational state. I propose to model these output-oriented---that
	is, \emph{action-oriented}---specifications along the lines utilized in the case of inputs. When focusing
	on input conditions, the schematic theory of representational content is the following: A state of an organism
	represents \emph{F}s if that state has the teleological function of being caused by \emph{F}s. I propose to
	add an additional set of conditions in which a state can come to represent \emph{F}s by allowing that a
	reversed direction of causation can suffice. A state of an organism represents \emph{F}s if that state
	has the teleological function of causing \emph{F}s.
\end{quotation}

Mandik then defines action-oriented representations (AOR) as any representation that has, in whole or in
part, imperative content. Mandik thus argues that active perception---instead of rejecting the representational
theory of perception---can contribute to the representational content of perception, and further, that
percepts themselves may sometimes be action-oriented representations~\citep{Mandik2005}.

It is evident from Mandik's argument that internal mental representations are necessary for perceiving and
understanding as well as interacting in the world. Further, it is obvious that these representations are
required to subsume a certain amount of perceptual experience and action knowledge allowing an agent to
plan for desired effects in the world. However, this still leaves us with our initial question of
\emph{what is an action}? What are the fundamental bits and pieces of both perceptual and sensori-motor
experience that require internal symbolization to account for a mental representation of an action \emph{A}?

Apart from Mandik, Jeannerod, in his famous book ``Motor Cognition: What the Body Tells the Self'' \citep{Jeannerod2006}
provides an alternate treatment of action representations. First of all, Jeannerod argues that action representations
must allow for mental simulation. Consequently, he distinguishes between \emph{covert} and \emph{overt} actions, where
the former are the mental representations and the latter the actual, overt movements. He thus immediately
attributes to action representations a functional nature~\citep{Vosgerau2009}, and hence argues
that representing and executing an action is functionally equivalent. Secondly, Jeannerod states that actions
are represented by their anticipated effect, that is, action representations essentially entail a mental model of
a needed future environmental state.~\cite{DeKleijn2014} further argue that such a representation in terms 
of an action's effects is unrenounceable as it unlocks contextualization of action control.
This submission immediately relates to Jeannerod's third characteristic criterion of actions which is related to 
the actual type of an action. Jeannerod submits that there are two types of actions, viz.\ conceptual and 
non-conceptual actions. The crucial difference is that action 
representations with a conceptual content require an explicit representation of the goal, whereas for non-conceptual actions 
the goal is readily present in front of the agent and the action can be executed automatically without an explicit 
internal representation of the goal. This difference crystallizes in Jeannerod's example of intending to call someone 
via a phone. The first part of this action is to grasp the handset which clearly requires an internal representation 
of the goal---the phone itself---prior to executing the action. At the time of the execution however, the 
representation loses its explicit character and the remaining action, i.e., dialing, is executed automatically.

Similar to Mandik's treatise, it is also evident from Jeannerod's work that actions are internally represented.
Contrarily to Mandik however, Jeannerod attributes to these representations a functional view by arguing that 
representing and executing an action is functionally equivalent. Whether one imagines or actually 
does an action employs the same neural substrates and processes~\citep{Jeannerod2006}. Jeannerod immediately provides
a clear distinction between the resulting types of actions, i.e., conceptual and non-conceptual, as well as their 
manifestation, overt and covert, viz.\ being actually executed or just imagined.

\subsection{Action in philosophy}

Independently of the discussions in psychology, philosophy---most notably Donald Davidson with his philosophy
of action---was looking for an answer to the question of \emph{what is an action}. Contrarily to CEC and REC however,
he aimed at identifying the relevant bits and pieces that physically constitute an action, independently of its mental
representation. According to Davidson, an action, in some basic sense, is something an agent does that was
\emph{intentional under some description}~\citep{Davidson2001}.
Davidson discusses this proposition in his famous example of someone accidentally alerting a burglar by 
illuminating a room, which she does by turning on a light, which she does by flipping the appropriate switch. 
Davidson is then concerned with the relation between the agent's act of turning on the light, her act of flipping 
the switch, etc., to answer the question which configuration of events, either prior to or contained 
within the extended causal process of turning on the light, really constitutes the agent's action. It is clear 
that there exists no unique answer to this question. Yet, the discussions caused by Davidson's example provide 
some insight into what may comprise an action. One may for example favor the overt arm movement that the 
agent performs, or the initiated causal process, but also the event of trying that precedes and ``generates'' the rest, 
i.e., the overt action. If for one second we stick to the latter definition of action, i.e., the mental act of 
trying, according to~\cite{OShaughnessy1973}, this implies \emph{willing}. Now according to 
O'Shaughnessy, an action then is defined as this mental act of willing which subsequently causes neural 
activity, muscle contractions and an overt actuation; happenings in the environment are just effects 
in the extended causal chain but not part of the action anymore. This however stands in stark contrast to 
De~Kleijn~\etal who submit that actions \emph{are events that unfold in time and that must be structured in 
such a way that their outcome satisfies current needs and goals}~\citep{DeKleijn2014}. Clearly, such a planned 
execution requires effects to chain the various deliberate events together.

\subsection{Action in neuroscience}\label{sec:2.4}

From a biological perspective, neuroscientists tried to link action with the neural substrates that generate it.
These studies belong to the more general research on the production of task-adapted serial behavior in human beings.
We summarize here the results from a roboticist's perspective but for in-depth studies on action representation and
neural substrates of motor control, see~\cite{GraftonAI2009} and~\cite{Hardwick2017} among others.

Researchers initially suggested that the hierarchy in information related to action (i.e.\ the goal constrains
the motor programs to be executed) was reflected by a hierarchical organization of the brain areas.~\cite{KeeleJ1992} 
used serial reaction time tasks in combination with attention to assess sequence learning.
Their results suggest that learning is easier when structure exists in the sequence, implying that the learnt
representation relies on the combination of elementary patterns ordered given the task, hence some hierarchy.

Grasping studies also highlighted the influence of abstract information on motor execution.~\cite{Jeannerod1984,Jeannerod1986} 
highlighted the interdependency between the formation of the grasp and the
reaching movement, the latter depending on the former, whereas~\cite{RosenbaumMV2001,RosenbaumVBJ1992} highlighted
how the the hand shape of the grasp depends on the geometry of the object, how the tool will be used and how
comfortable is the final posture.

Computational models have included action representation with both explicit~\citep{CooperS2006} and emergent hierarchy~\citep{Botvinick2008}
and successfully explained behavioral results. However, these models stayed at a representational level and did not directly adress
the question of which neural substrates support the representation of action itself.
A first proposition by~\cite{Fuster1995} tried to map anatomy with the expected hierarchy in the action representation.
Imaging studies \citep{RolandLLS1980,RolandSLL1980} showed that motor cortex is only active during real movement execution
whereas the supplementary motor area (SMA) is active during both executed and imagined movement. These results were interpreted
as a sign that motor cortex and SMA play a role at different levels of abstraction and thus support the anatomical/functional hierarchy hypothesis.

However, several arguments come in opposition of a direct mapping between anatomy and functional hierarchy. We focus here on two of the four
developed by \citet[p.~643]{GraftonAI2009}. First, a hierarchical model assumes a clear separation between the different levels and
that only the lowest level is in charge of producing movement. However, it has been shown that even higher-level areas
(premotor and parietal cortex, extrapyramidal brain stem pathways) project to the spinal cord and thus
potentially influence the movement \citep{DumS1991, DumS1996}. Secondly, the conceptual implication of a strict anatomical hierarchy raises
the problem of the homonculus: if there is a decisional component on top of the architecture, this component itself may be organized
hierarchically including a decisional component, etc. The resulting model would be complex which does not fit with the results
on how fast and adaptable the action decision-making process actually is~\citep{DesmurgetG2000}.

More recent studies of the anatomy have highlighted the existence of multiple parallel parietal-premotor-prefrontal loops in the brain.
These loops seem to integrate multimodal sensory information rather than being tied to one modality only. They have been associated
with object-centered action, tool use and reaching~\citep{JohnsonG2003,RizzolattiL2001, RizzolattiM2003}. Grafton~\etal suggest that 
the hierarchy of action representations is thus not tied to the anatomy itself but rather that~\citep[p.~643]{GraftonAI2009}

\begin{quotation}
	[\dots] an anatomical organization with multiple parallel parietal-prefrontal and premotor pathways supports a
	multitude of relative hierarchies that can be flexibly recruited as a function of task demands, experience, and context.
	In this framework, there are dissociable functional anatomic substrates, but these are not constrained by a fixed hierarchy.
	This shifts the focus of inquiry to understanding representational hierarchies that are highly flexible and goal based.
\end{quotation}

This second hypothesis has been investigated by focusing on the goal representation in motor execution studies involving grasping and 
bimanual coordination tasks. Grasping tasks directly map the goal to the target object, thus the task can be reframed as the 
problem of finding the proper transform between the perceived object and the hand. The anterior IntraParietal Sulcus (aIPS) in the 
parietal cortex has been shown to be critical for computing these sensorimotor transformations. The problem is then how the transformation 
information and goal representation are merged, that is, how does the aIPS perform the sensorimotor integration of the information?

Due to its connectivity to aIPS, the ventral premotor cortex is supposed to hold the goal representation. The hierarchical anatomy hypothesis would
suggest that the sensorimotor information related to the target object is transformed into a goal representation. However, the hypothesis
of a flexible hierarchy suggests that aIPS merges the sensorimotor and goal information and produces the constraints on the motor commands.
This is supported by transcranial magnetic stimulation (TMS) studies~\citep{TunikFG2005}. Tunik~\etal studied reaching and
grasping tasks where the target object orientation (thus the goal) was changed very fast. The TMS was shown to disturb
the ability of subjects to adapt to changes of the goal. The TMS blocks not only the adaptation of the grasp aperture but
also the arm orientation. The authors claim that these results are better explained by the fact that aIPS
does sensorimotor integration of the goal information rather than that TMS disrupts lower motor processes such as grip aperture.
Consistent results are found in bimanual coordination: the change in the task goal changes the amplitude of
the neural activity but does not change which regions are activated. Hence, there are areas (ventral premotor cortex
and anterior intraparietal sulcus) in charge of maintaining the goal information, consistently recruited over
tasks, that, when disturbed, have an effect on the adaptation of movement.

A similar dichotomy is shown in action observation tasks: Using the fMRI adaptation phenomenon (Repetition Suppression or RS), 
\cite{HamiltonG2006} were able to show that the left aIPS is sensitive to which object is grasped (thus the \enquote{goal} of the action)
whereas the information on the object position produces RS in other parts of the brain. They interpret this double dissociation as
a result in favor of hierarchy between the goal of the action and the kinematic information of the action. In further studies,
they manipulated the shape of the grasp~\citep{HamiltonG2007} or the outcome of actions~\citep{HamiltonG2008} and were able to
highlight segregated RS effects in specific areas of the brain.
In the end, they argue that~\cite[p.~648]{GraftonAI2009}

\begin{quotation}
	[\dots] together, these three experiments support a model of representational hierarchy that distinguishes action means, kinematics,
	object-centered behavior, and ultimately, action consequences. The decoding of object-centered action appears to be strongly left
	lateralized, whereas the decoding of more complex action intentions arising as a consequence of the action engaged bilateral frontal-parietal circuits.
\end{quotation}

Actions are thus not uniquely represented in the brain but the representation is rather generated by the recruitment of several areas, 
with an apparent distinction between the goal-level information and the motor-related information. Moreover,~\cite{Hardwick2017} recently
did a meta-analysis on more than a thousand works from the literature on motor imagery (the mental rehearsal of an action), action observation 
(observing others' action execution) and movement execution (the overt interaction in the environment). They identified a consistent recruitment 
of a network of cortical or subcortical regions for each function. Both motor imagery and movement execution recruit the putamen which is 
involved in movement regulation. The body representation, encoded by the cerebellum, is also involved in motor imagery and movement execution 
along with the anterior and posterior midcingulate cortex for motor control. Action observation however does not recruit subcortical structures. It recruits the 
premotor parietal and occipital regions but less than during motor imagery.

These results from biology should teach roboticists two main lessons:
\begin{itemize}
	\item The outcome of an action is a crucial part that defines it. There are dedicated areas to encode the goal
	and use the goal information to constrain the movement. Thus, an action in robotics should be defined by the goal
	it is intended to achieve, that is, its expected effects. The production of movement is then adapted to this goal.
	Thus, robot controllers should be flexible rather than reproduce stereotypical motions.
	\item Action requires multiple types of information that are not encoded in a central representation but rather
	distributed over and shared among multiple brain areas depending on the functional goal. For robotics, this argues
	in favor of a flexible representation of an action that links goal, movement and the currently-perceived scene.
\end{itemize}

Summarizing the above discussion clearly shows that despite being a core aspect of mammalian behavior, 
today we still lack a precise answer to the question of \emph{what is an action}. Yet, this discussion however 
also shows that actions (i) are internally represented (\emph{c.f.}~\citealt{RizzolattiL2001,Rizzolatti2004}),
(ii) are tightly bound to perception as a genuine source of information for action selection \citep{TunikFG2005}, and (iii) yield 
effects which play a crucial role in shaping one's behavior \citep{HamiltonG2008}.


\subsection{Action in robotics}\label{sec:2.1}

The notion of action occupies a paramount role in robotics. This simply stems from the
circumstance that in order to meaningfully and intentionally interact with the world a
robot requires knowledge about when to apply a specific action in order to achieve desired
effects in the world. As Newton writes in her recent work on understanding and 
self-organization~\cite[p.~5]{Newton2017},
\begin{quotation}
	Understanding is tightly coupled with the need of a living organism to take 
	action. Understanding involves knowing how we might perform goal-directed actions 
	relative to the environment. The experience of understanding is a feeling that 
	the action affordances of a situation are not entirely unclear. Action (as opposed 
	to reaction) requires imagery, including motor imagery, that can be used in the 
	guidance of action.
\end{quotation}
Clearly, appropriate action representations are thus paramount for bootstrapping the 
development of an understanding of the world and ways an autonomous agent can meaningfully 
interact with this very world.

This paramount role of action representations was already pointed out by~\cite{Kruger2007}.
In their survey they discuss the meaning of action at different levels in robotics from
plain low-level sensory observations to high-level cognitive recognition and planning tasks.
Kr\"uger \etal argue that in order to nail down the meaning of action in robotics needs
to address several areas, viz.\ observing and imitating others, control of one's own body,
and learning of affordances~\citep{Zech2017}. Their subsequent discussion provides an initial
but yet unsatisfying answer to \emph{what is an action}. However, we can clearly see that 
perception, embodiment, actuation and goal representation are core aspects of actions. We
thus conjecture that such information requires a representation in order to be recallable.
On the other hand it is necessary to talk about representations in the context of robotics
as symbolic information, i.e., representations of knowledge, is crucial for computation. Aligned with
the above discussion, we propose the following seminal definition of the notion of an action 
from a roboticist's stance in the next section.

\subsection{A Seminal definition of action from a roboticist's stance}\label{sec:2.2}

Motivated by the discussions so far we define the notion of an action for robotics as
\begin{itemize}
	\item something an agent does that was intentional under some description,
	\item is caused by both the agent's current internal state and external percepts,
	\item is adaptive and deterministic to achieve desired effects,
	\item is learnt and symbolized while observing and imitating other agents, 
	\item is mechanically effective,
	\item and primarily represented by its anticipated effects, that is, the goal.
\end{itemize}
Clearly, this definition is not final. However, we claim that it provides an initial basis
for discussing what information, and especially in which form, eventually is required in
order to elicit a general representation of actions for robots. It is obvious that perceptual
aspects play a crucial role by virtue of the mutual relationship between perception and
action~\citep{Bamert2009}. Further, learning plays an important role. Analogously to human
development, one of the long-term goals in robotics research is to equip agents with robust 
learning capabilities about their environment and their own embodiment. Learning new means 
to interact with the environment, i.e., new actions, is paramount as not all situations an 
autonomous agent will experience are predictable. Thus, whereas providing initial knowledge 
about action bootstraps an agent's autonomy, the capability to adapt motions related to actions 
and subsequently learn new actions from experience is necessary to allow the agent to achieve 
novel effects that go beyond its current experience. As highlighted in Section~\ref{sec:2.4}, 
this can be achieved by integrating observations and experience from early sensory areas to 
higher-order cortical areas (\emph{c.f.}~\citealt{Hasson2015}).

Another important aspect of actions is their mechanical effectivity by causing overt 
changes in the environmental state; lacking a mechanically effective nature reduces an action 
to a mere gesture~\citep{Hobaiter2017}. Last but not least, actions---at least in the context 
of robotics---require external information that can be symbolized internally for goal-driven, 
behavioral planning. As already pointed out by \cite{Steels2003}, action representations 
are inevitable for planning. Given this seminal definition, in the next section we introduce 
our taxonomy for action representations in robotics.


\section{Classification criteria for action representations}\label{sec:3}

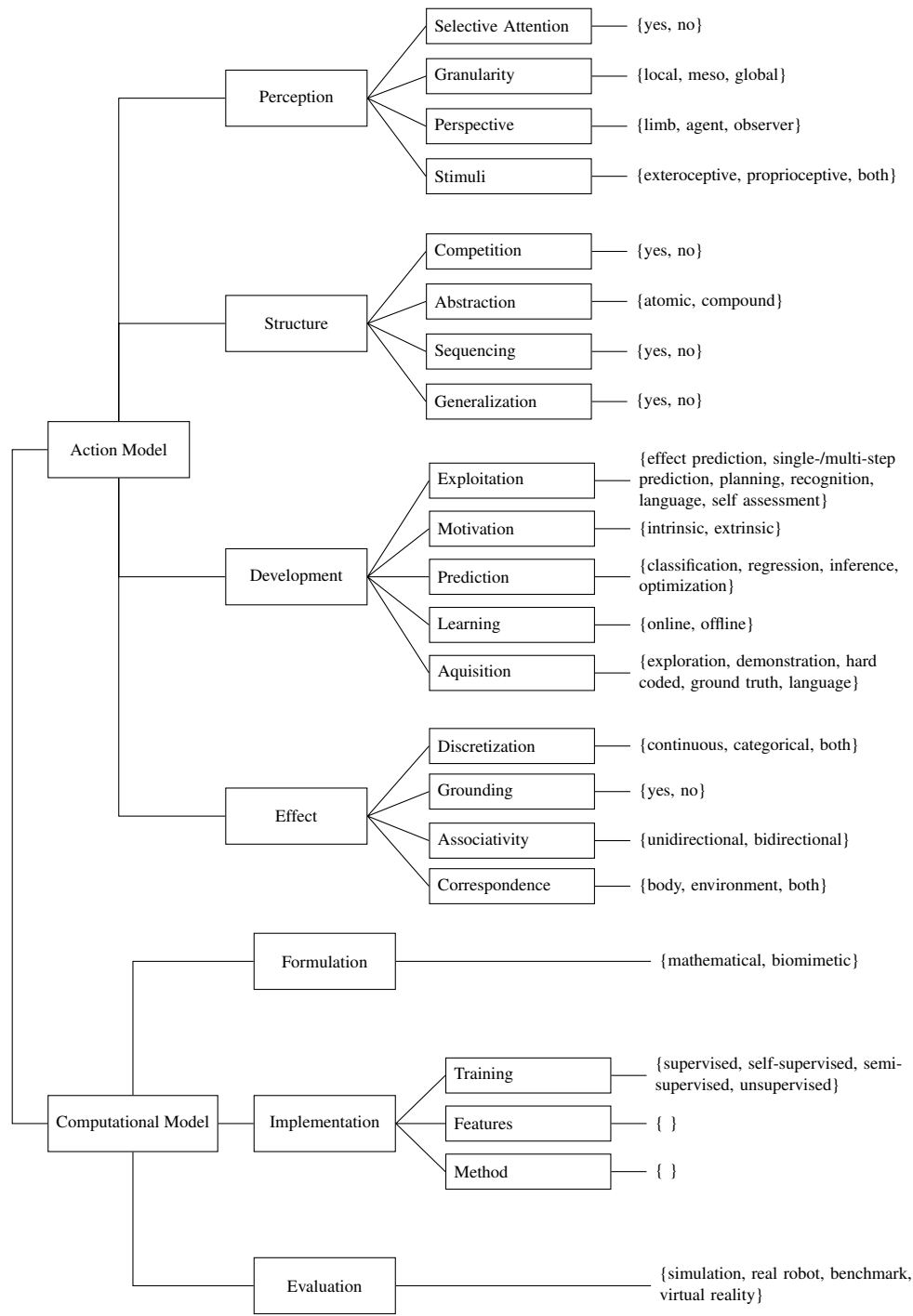
\begin{figure*}[htb]
	\centering
	\begin{center}
\begin{tikzpicture}[>=stealth]
  \node (root) at (0,0) [anchor=west, minimum height=0.8cm] {};
  \begin{scope}[every node/.style={draw=black, minimum width=2cm, minimum height=0.8cm, font=\scriptsize, align=center}]    

    \node (AM) at (root.north) [ yshift=1.8cm, xshift=0.5cm, anchor=west] {Action Model};
    \node (Representation) at (AM.east)  [ anchor=west, xshift=0.5cm, yshift=1.8cm] {Structure};
    \node (Perception) at (Representation.north)  [ anchor=south, yshift=2.4cm ] {Perception};
    \node (Development) at (Representation.south)  [ anchor=north, yshift=-2.8cm] {Development};
    \node (Effect) at (Development.south)  [ anchor=north, yshift=-2.6cm] {Effect};

    \node (CM) at (root.south) [ yshift=-7cm, xshift=0.5cm, anchor=west ] {Computational Model};
    \node (Implementation) at (CM.east) [ xshift=0.5cm, anchor=west ] {Implementation};
    \node (Formulation) at (Implementation.north) [ anchor=south, yshift=1.5cm ] {Formulation};
    \node (Evaluation) at (Implementation.south) [ anchor=north, yshift=-1.5cm] {Evaluation};

    \draw (root.center) |- (AM.west) 
            (AM.north)  |- (Representation.west)
            (AM.north) |- (Perception.west)
            (AM.south) |- (Development.west)
            (AM.south) |- (Effect.west)
          (root.center) |- (CM.west) 
            (CM.east)  |- (Implementation.west)
            (CM.north) |- (Formulation.west)
            (CM.south) |- (Evaluation.west)
    ;
  \end{scope}
  \begin{scope}[every node/.style={draw=black, minimum height=0.5cm, text width=2.1cm, font=\scriptsize, align=left, anchor=north, fill=none }]
    \node[above right=0.85cm and 2cm of Perception, anchor=south, yshift=-1.2cm] (Granularity) {Granularity};
    \node[above=0.2cm of Granularity] (SelectiveAttention) {Selective Attention}; 
    \node[below right=0.85cm and 2cm of Perception, anchor=south, yshift=0.6cm] (Perspective) {Perspective};
    \node[below=0.2cm of Perspective] (Stimuli) {Stimuli};
  \end{scope}
  \draw (Perception.east) -- (Stimuli.west) 
        (Perception.east) -- (Perspective.west)
        (Perception.east) -- (Granularity.west)
        (Perception.east) -- (SelectiveAttention.west)
 ; 
  \begin{scope}[every node/.style={draw=black, minimum height=0.5cm, text width=2.1cm, font=\scriptsize, align=left, anchor=north, fill=none }]
    \node[above right=0.85cm and 2cm of Representation, anchor=south, yshift=-1.2cm] (Abstraction) {Abstraction};
    \node[above=0.2cm of Abstraction] (Competition) {Competition}; 
    \node[below right=0.85cm and 2cm of Representation, anchor=south, yshift=0.6cm] (Sequencing) {Sequencing};
    \node[below=0.2cm of Sequencing] (Generalization) {Generalization};
  \end{scope}
  \draw (Representation.east) -- (Abstraction.west) 
        (Representation.east) -- (Competition.west)
        (Representation.east) -- (Sequencing.west)
        (Representation.east) -- (Generalization.west)
 ; 

  \begin{scope}[every node/.style={draw=black, minimum height=0.5cm, text width=2.1cm, font=\scriptsize, align=left, fill=none }]
    \node (Prediction) at (Development.east)  [xshift=0.87cm, anchor=west] {Prediction};
    \node (Learning) at (Prediction.north) [anchor=south, yshift=-1.2cm] {Learning};
    \node (Aquisition) at (Learning.north) [anchor=south, yshift=-1.2cm] {Aquisition};
    \node (Motivation) at (Prediction.south) [anchor=north, yshift=1.2cm] {Motivation};
    \node (Exploitation) at (Motivation.south) [anchor=north, yshift=1.2cm] {Exploitation};
  \end{scope}
  \draw (Development.east) -- (Prediction.west) 
        (Development.east) -- (Aquisition.west)
        (Development.east) -- (Motivation.west)
        (Development.east) -- (Exploitation.west)
        (Development.east) -- (Learning.west)
  ;

  \begin{scope}[every node/.style={draw=black, minimum height=0.5cm, text width=2.1cm, font=\scriptsize, align=left, fill=none }]
    \node (Discretization) at (Effect.east)  [xshift=0.87cm, yshift=1.0cm, anchor=west] {Discretization};
    \node (Grounding) at (Effect.east)  [xshift=0.87cm, yshift=0.35cm, anchor=west] {Grounding};
    \node (Associativity) at (Effect.east)  [xshift=0.87cm, yshift=-0.35cm, anchor=west] {Associativity};
    \node (Correspondence) at (Effect.east)  [xshift=0.87cm, yshift=-1.0cm, anchor=west] {Correspondence};
  \end{scope}
  \draw (Effect.east) -- (Discretization.west) 
        (Effect.east) -- (Grounding.west)
        (Effect.east) -- (Associativity.west)
        (Effect.east) -- (Correspondence.west)
  ;

  \begin{scope}[every node/.style={draw=black, minimum height=0.5cm, text width=2.1cm, font=\scriptsize, align=left, fill=none }]
    \node (Features) at (Implementation.east)  [xshift=0.7cm, anchor=west] {Features};
    \node (Method) at (Features.north) [anchor=south, yshift=-1.2cm] {Method};
    \node (Training) at (Features.south) [anchor=north, yshift=1.2cm] {Training};
  \end{scope}
  \draw (Implementation.east) -- (Features.west) 
        (Implementation.east) -- (Method.west)
        (Implementation.east) -- (Training.west)
  ;

  \begin{scope}[every node/.style={minimum height=0.5cm, text width=3.7cm, font=\scriptsize, align=left, anchor=north, fill=none }]
    \node (selat_choice) at (SelectiveAttention.east) [anchor=west, xshift=0.5cm] {\{yes, no\} };
    \node (stim_choice) at (Stimuli.east) [anchor=west, xshift=0.5cm] {\{exteroceptive, proprioceptive, both\} };
    \node (gran_choice) at (Granularity.east) [anchor=west, xshift=0.5cm] {\{local, meso, global\} };
    \node (pers_choice) at (Perspective.east) [anchor=west, xshift=0.5cm] {\{limb, agent, observer\} };

    \node (abst_choice) at (Abstraction.east) [anchor=west, xshift=0.5cm] {\{atomic, compound\} };
    \node (comp_choice) at (Competition.east) [anchor=west, xshift=0.5cm] {\{yes, no\} };
    \node (seq_choice) at (Sequencing.east) [anchor=west, xshift=0.5cm] {\{yes, no\} };
    \node (gen_choice) at (Generalization.east) [anchor=west, xshift=0.5cm] {\{yes, no\} };

    \node (aq_choice) at (Aquisition.east) [anchor=west, xshift=0.5cm] {\{exploration, demonstration, hard coded, ground truth, language\} };
    \node (pred_choice) at (Prediction.east) [anchor=west, xshift=0.5cm] {\{classification, regression, inference, optimization\} };
    \node (expl_choice) at (Exploitation.east) [anchor=west, xshift=0.5cm] {\{effect prediction, single-/multi-step prediction, planning, recognition, language, self assessment\} };
    \node (learn_choice) at (Learning.east) [anchor=west, xshift=0.5cm] {\{online, offline\} };
    \node (mot_choice) at (Motivation.east) [anchor=west, xshift=0.5cm] {\{intrinsic, extrinsic\} };

    \node (disc_choice) at (Discretization.east) [anchor=west, xshift=0.5cm] {\{continuous, categorical, both\} };
    \node (ground_choice) at (Grounding.east) [anchor=west, xshift=0.5cm] {\{yes, no\} };
    \node (assoc_choice) at (Associativity.east) [anchor=west, xshift=0.5cm] {\{unidirectional, bidirectional\} };    
    \node (corresp_choice) at (Correspondence.east) [anchor=west, xshift=0.5cm] {\{body, environment, both\} };    

    \node (form_choice) at (Formulation.east) [anchor=west, xshift=3.61cm] {\{mathematical, biomimetic\} };
    \node (train_choice) at (Training.east) [anchor=west, xshift=0.5cm] {\{supervised, self-supervised, semi-supervised, unsupervised\} };
    \node (eval_choice) at (Evaluation.east) [anchor=west, xshift=3.61cm] {\{simulation, real robot, benchmark, virtual reality\} };
    \node (feat_choice) at (Features.east) [anchor=west, xshift=0.5cm] {\{$~$\} };
    \node (method_choice) at (Method.east) [anchor=west, xshift=0.5cm] {\{$~$\} };
  \end{scope}
  \draw (SelectiveAttention.east) |- (selat_choice.west) 
        (Evaluation.east) |- (eval_choice.west) 
        (Granularity.east) |- (gran_choice.west) 
        (Stimuli.east) |- (stim_choice.west) 
        (Perspective.east) |- (pers_choice.west) 
        (Abstraction.east) |- (abst_choice.west) 
        (Competition.east) |- (comp_choice.west) 
        (Sequencing.east) |- (seq_choice.west) 
        (Aquisition.east) |- (aq_choice.west) 
        (Prediction.east) |- (pred_choice.west) 
        (Generalization.east) |- (gen_choice.west) 
        (Exploitation.east) |- (expl_choice.west) 
        (Learning.east) |- (learn_choice.west) 
        (Formulation.east) |- (form_choice.west) 
        (Training.east) |- (train_choice.west) 
        (Method.east) |- (method_choice.west) 
        (Features.east) |- (feat_choice.west)  
        (Motivation.east) |- (mot_choice.west)  
        (Discretization.east) |- (disc_choice.west)  
        (Grounding.east) |- (ground_choice.west)  
        (Associativity.east) |- (assoc_choice.west)
        (Correspondence.east) |- (corresp_choice.west)
  ;
\end{tikzpicture}

\end{center}
	\caption{Overview of our taxonomy for categorizing action representations
	in robotics. For the sake of clarity, the choice \emph{not specified}
	is excluded.}\label{fig:taxonomy}
\end{figure*}

Given our discussions from Section~\emph{\ref{sec:2}} we can now introduce our taxonomy and 
its classification criteria for action representations in robotics. Clearly, a sound notion 
of action is paramount in that its representation for a robot is successful. Motivated by this 
we thus define an action representation in robotics as the union of an underlying \emph{action} 
model and a \emph{computational} model. Consequently, the action model deals with perceptual, 
structural, developmental and effect-related aspects, that is, the nature and embodiment of actions. 
In contrast, the computational model addresses low-level, implementational aspects of the mechanics 
of actions. Figure~\ref{fig:taxonomy} gives an overview of our taxonomy and its classification 
criteria.

Before now discussing the criteria from Figure~\ref{fig:taxonomy} in detail in Sections \emph{\ref{sec:3.1}} 
and \emph{\ref{sec:3.2}}, we want to remark that if a criterion is not specifically addressed in a given 
publication, it is assigned \emph{not specified}.

\subsection{Action model criteria}\label{sec:3.1}

Action model criteria serve to asses the underlying ``mental'' action
model of an action representation regarding its perceptual, structural,
developmental, and effect-related aspects.

\subsubsection{Perception}\label{sec:3.1.1}

Perceptual aspects study the means by which an autonomous agent employs different
aspects of perceptual input for recognizing and memorizing actions in the environment.
This dimension is standing in reason when considering Mandik's claim that perception
and action are tightly coupled~\citep{Mandik2005}. An even stronger argument towards
this tight linkage is given by \cite{Tucker1998} in arguing that
\emph{seen objects automatically potentiate components of the actions they afford}.
Thus, one should consider visual inputs as one of the main drivers demarcating representations
of actions.

\paragraph{Selective Attention}\label{sec:3.1.1.3}

Selective attention is becoming more and more popular in vision research, not least because
of the impressive success of Deep Q-Learning~\citep{Sorokin2015}. Naturally, selective attention
is an important process for early action selection~\citep{Cisek2010}. Further, it allows
noise and irrelevant information to be filtered out, focusing on what is important and relevant, thus raising
awareness of one's own actions and ultimately culminating in conscious motor control~\citep{Webb2016}.
Thus, selective attention is either present or not (see
rows 1 or 7--11, and 2--6 or 16--39, respectively, of Table~\ref{tab:class_cl}).

\paragraph{Granularity}\label{sec:3.1.1.4} 

The granularity of the perceptual aspects of an
action are important when it comes to generalizing actions. Clearly, in the context of a
scene, actions can be perceived at different levels of granularity:
\begin{itemize}
	\item \emph{local} implies that an action model only considers local information, i.e.,
		the part of an object that is relevant for doing the action like the handle of a hammer.
		As in the case of the perspective (\emph{c.f.}~Section~\ref{sec:3.1.1.1}) this comes with
		both advantages and disadvantages. For example, the agent may be capable of immediate interaction
		with the object upon recognizing a part but may fail to generalize its knowledge to different
		situations due to the lack of additional semantic information regarding the context in 
		which the action is performed (see rows 34, 69 or 72 of Table~\ref{tab:class_cl}).
	\item \emph{meso} implies that an agents perceives an action at the level of complete objects instead
		of only specific parts. This immediately allows an agent to acquire additional semantic information
		on the object itself enabling easier generalization of an action to different contexts as the agent
		has a more elaborate idea of what it can and cannot do with an object (see rows 1--2 or 35--38 of Table~\ref{tab:class_cl}).
	\item \emph{global} implies that an agent perceives an action at the scene level. That is, not only
		does it perceive the concrete movements and objects involved but is also able to perceive the
		environmental context in which the action is performed, thus enabling consideration of interactions 
		in the environment. Clearly, this allows an agent to easily generalize actions to novel contexts 
		as it has acquired a complete picture of the circumstances under which an action can be performed. 
		Observe however that this level of granularity does not readily imply generalization of the action 
		(\emph{c.f.}~Section~\ref{sec:3.1.2.4}; see rows 3--4, 9 or 11--12 of Table~\ref{tab:class_cl}).
\end{itemize}

\paragraph{Perspective}\label{sec:3.1.1.1} 

The perspective eventually nails down the
reference frame of the perceived action. In the case of autonomous agents, multiple perspectives
may apply given how the agent perceives and memorizes an action. We claim that there are
three relevant perspectives autonomous agents can employ:
\begin{itemize}
	\item \emph{limb} implies that an agent learns actions with respect to one of its limbs,
		e.g., an arm or the end-effector only. The rationale is that our limbs are the primary 
		means of interaction with the environment. This perspective has the advantage that an 
		agent may easily plan and adapt its actions locally, however may fail to do so at a 
		global scale (see Section~\ref{sec:3.1.1.4}). Observe that this choice may imply the need 
		for selective attention to properly isolate observations (see Section~\ref{sec:3.1.1.3}; see rows 27, 38 or 42 of Table~\ref{tab:class_cl}).
	\item \emph{agent} implies that an agent perceives actions with reference to its whole body.
		This clearly has the advantage that an agent is able to plan and redo actions at a scale
		relevant for his body, yet it may fail to capture fine-grained local aspects of an action. 
		Compared to \emph{limb} this choice usually refers to whole-body actions (see rows 2 or 5--6 of Table~\ref{tab:class_cl}).
	\item \emph{observer} implies that an agent learns actions by observing them and associating
		them to the frame of reference of the agent executing the action, e.g., agents perceive
		actions from a third-person perspective. Clearly, the resulting action is represented at
		a global scale, yet the agent is required to---prior to execution---map the action into
		its own reference frame (see rows 22--24 or 30--32 of Table~\ref{tab:class_cl}).
\end{itemize}

\paragraph{Stimuli}\label{sec:3.1.1.2}

Stimuli, either external or internal, play an important role for action learning and representation
as they encode relevant information that (i) triggers, (ii) monitors, (iii) allows adaption of an 
action both prior and during execution. Clearly, such stimuli may have different sources, e.g., 
internal or external. This criterion thus considers two types of stimuli:
\begin{itemize}
	\item \emph{proprioceptive} stimuli which relate to stimuli that are produced within the
		agent and its embodiment, e.g., force readings. Such stimuli are essential in that 
		they enable monitoring the self during action execution (see rows 72--73 or 83 of Table~\ref{tab:class_cl}).
	\item \emph{exteroceptive} stimuli which relate to stimuli that are generated in the external 
		environment, i.e., interaction possibilities in the environment (affordances). Such 
		stimuli are necessary for an agent to perceive the effects of its actions in the
		environment and subsequently replan or perform online adaptation of its movements  
		to achieve its intended goals (see rows 30--38 or 40--60 of Table~\ref{tab:class_cl}).
\end{itemize}
Observe that this is a multi-choice criterion, i.e., an agent may as well consider both
proprioceptive and exteroceptive stimuli for establishing an action model (see rows 2 or 5--7 of Table~\ref{tab:class_cl}).

\subsubsection{Structure}\label{sec:3.1.2}

Structural aspects of the action model discuss the capacities of
the representation in terms of cognitive capabilities it opens up to an agent. They are crucial
for planning and reasoning for action selection in any given context. From an environmental
perspective, structural aspects additionally discuss how the actions are organized in the
environment.

\paragraph{Competition}\label{sec:3.1.2.2}

Obviously there may not always exist a single action that achieves an intended
effect but instead a variety of actions equally allowing an agent to reach its goal, i.e., 
multiple actions are equivalent in terms of their effects but differ in their overt manifestation. 
To be able to select the ideal action, an action model is thus required to allow for competition 
among actions such that the agent may always choose the most suitable and efficient action.
However, we do not attempt to study the internals of action competition but rather whether
a model allows for it or not. Thus, Competition is either present or not.
(see rows 1--7 or 10--14, and 8--9, 15--16 or 18--19, respectively, of Table~\ref{tab:class_cl}).

\paragraph{Abstraction}\label{sec:3.1.2.1}

Traditionally an action is considered atomic by triggering a specific movement
applied in a specific context to achieve an intentional effect. However, considering actions
only at such an atomic level subsequently hinders an agent to plan in terms of action
sequences composed of a set of atomic actions. Our taxonomy thus considers both of these
levels of abstraction as this ultimately enables an agent to reason in terms of higher-level
actions and their goals:
\begin{itemize}
	\item \emph{atomic} actions encapsulate a single intentional effect. Atomic at this implies
		that an action cannot be further decomposed into smaller actions. Observe however that this
		does not restrict an atomic action to consist of a series of movements.  For example, opening a drawer
		requires placing the gripper by moving the arm towards it, closing the hand around the handle,
		and subsequently retracting the arm (see rows 1--7 or 9--22 of Table~\ref{tab:class_cl}).
	\item \emph{compound} actions on the contrary are actions that themselves consist of multiple
		atomic actions. That is, compound actions describe sequences of actions where
		these actions are combined and conditioned on their intermediary, intentional effects.
		Similarly to atomic actions, the agent usually aims at achieving again 
		a single intended effect, yet at a larger timescale (see rows 23, 59--60 or 63 of Table~\ref{tab:class_cl}).
\end{itemize}
Observe that this is a multi-choice criterion, i.e., an agent may as well consider both
atomic and compound actions when building its internal repertoire of action models (see rows 8, 58 or 100 of Table~\ref{tab:class_cl}).


\paragraph{Sequencing}\label{sec:3.1.2.3}

Being able to sequence actions eventually allows an agent to join both atomic and
compound actions to reason about higher-level action goals and to
achieve a variety of intended effects. Yet, we want to clarify that sequencing of actions
does not readily imply that an agent is able to represent compound actions (see Section~\ref{sec:3.1.2.1}).
Sequencing solely refers to the ability to generate long-term plans that
may yield a variety of effects. Further, this criterion by no means studies the means of
sequencing. Thus, sequencing is either present or not (see rows 1, 5 or 39--40, and 2--4 or 29--34, 
respectively, of Table~\ref{tab:class_cl}).

\paragraph{Generalization}\label{sec:3.1.2.4}

One of the most crucial aspects of autonomous robots is the capacity to 
generalize acquired knowledge to novel situations. Clearly, such a capacity places demands on the action representations. What would be the benefit of learning an action if it 
cannot be generalized to novel situations? Our taxonomy thus also studies this aspect of 
action representations as it holds a crucial factor for the success of an action representation.
Again, however, we are not interested in the actual means of generalization at a computational level
but just in whether the model allows it or not.Thus, generalization is either present or not 
(see rows 1--21 or 23--60, and 84, 104 or 132, respectively, of Table~\ref{tab:class_cl}).

\subsubsection{Development}\label{sec:3.1.3}

Developmental aspects of an action relate to the means by which an agent is able to process
new information to extend its action knowledge. Observe that this dimension is strongly tied
to the perceptual aspects (see Section~\ref{sec:3.1.1}) of the action model in that the
percepts ultimately constrain what can be learned. However, contrary to perceptual aspects 
which study \emph{how the agent perceives} the environment for interacting with it, developmental 
aspects study \emph{how the agents learns to interact} with its environment.

\paragraph{Exploitation}\label{sec:3.1.3.4}

Available action knowledge can be exploited in different ways. However, different ways of exploiting 
one's knowledge result in different ways of how one subsequently interacts with the environment. Over 
the last decades roboticists have studied different ways of exploiting action knowledge where the range 
varies from selecting actions for reactive behavior to reasoning about actions for higher-level cognition:
\begin{itemize}
	\item \emph{effect prediction} of actions is an important capacity
		for autonomous agents as it allows them to understand both their environment but also
		their embodiment in terms of what they are capable of achieving. Additionally, effect 
		prediction is a precursor for planning at large timescales (see rows 25, 64 or 76 of Table~\ref{tab:class_cl}).
	\item \emph{single-/multi-step prediction} enables agents on the grounds of their immediate percepts
		and motivation to first search applicable actions and subsequently sequence them together
		given the predicted effects, or just to execute the most suitable action (see rows 1--2, 6 or 9--12 of Table~\ref{tab:class_cl}).
	\item \emph{planning}, in contrast to single/multi-step prediction, cannot be done by
		exhaustive search. Rather, planning is implemented by reasoning over symbolic representations
		of both the environment and the agent's percepts and motivation, as well as its internally-symbolized action repertoire (see rows 13, 19 or 22 of Table~\ref{tab:class_cl}).
	\item \emph{recognition} of actions and activities of others is crucial for
		autonomous agents that are supposed to help in our daily lives. Observe that this choice
		relates to effect prediction, yet at a different level. Whereas effect prediction ultimately
		allows an agent to predict what was the intention, action recognition allows an agent to
		already reason about how to achieve the intended goal instead of just capturing the sole
		intention (see rows 3--4, 7--8 or 16--18 of Table~\ref{tab:class_cl}).
	\item \emph{language} enables agents to communicate with other agents by an important high-level 
		cognitive ability. Agents exploiting their action knowledge by language ultimately are 
		capable of communicating this knowledge in order to instruct others by 
		means of teaching. Similarly, agents can also learn from spoken instructions (see Section~\ref{sec:3.1.3.2}; see row 52 of Table~\ref{tab:class_cl}).
	\item \emph{self-assessment} of one's own capabilities unlocks to an autonomous agent the 
		possibility of reasoning about its developmental state. This readily aligns with Jeannerod's 
		famous idea that our actions tell us about ourselves \citep{Jeannerod2006}. Further, being 
		able to assess one's self and one's capacities and consequently knowledge gaps immediately 
		allows one to tackle the exploitation vs.\ exploration trade-off by improving learned or 
		acquiring new knowledge (\emph{c.f.}~Section~\ref{sec:3.1.3.1}).
\end{itemize}

\paragraph{Motivation}\label{sec:3.1.3.1}

Clearly an agent needs some kind of motivation that drives its process of knowledge acquisition.
Such a motivation may either be external or internal. The former relates to external triggers, 
usually externally-imposed goals the robot is to achieve. The latter refers to internal 
motivations with no separable (clearly observable) outcome by an instrumental value~\citep{Ryan2000}. 
Consequently, this criterion has two possible choices:
\begin{itemize}
	\item \emph{extrinsic} motivation generally relates to external triggers that drive a robot to 
		acquire new action knowledge. 
		Observe that such extrinsic motivations may at some point overlap with intrinsic motivation 
		(see below) in the case that an agent ``realizes''---despite being externally imposed---that following 
		some trigger may result in an overall improvement. In such an event we argue similarly to
		Ryan~\&~Deci that this still should be considered external, as the original trigger is
		externally imposed (\citealt{Ryan2000}; see rows 1, 26 or 33 of Table~\ref{tab:class_cl}).
	\item \emph{intrinsic} motivation relates to internal triggers that drive the robot towards
		fostering or acquiring novel actions. The difficulty arising here is that robots generally 
		are not able to deal with non-separable consequences like joy or satisfaction, which commonly
		are considered as triggers for intrinsically-motivated behavior~\citep{Ryan2000}. Yet, discussing 
		this question is not the goal of our work, which is why we deliberately leave this question 
		unanswered. Apart from that, intrinsic motivation has the disadvantage that the robot has
		to confront the exploration vs.\ exploitation trade-off, i.e., does it learn new actions
		or foster existing actions? On the contrary however, being intrinsically motivated enables
		an agent to learn what it is capable of and thus to develop an understanding of its embodiment (see rows 9, 106 or 152 of Table~\ref{tab:class_cl}).
\end{itemize}
Observe that this is a multi-choice criterion, i.e., an agent may be both extrinsically and
intrinsically motivated in learning new actions.

\paragraph{Prediction}\label{sec:3.1.3.3}

After having learned new actions an agent needs the capacity to predict when a certain
action is applicable (or required) given both its percepts and its motivation. Obviously,
this criterion has a strong relation to the underlying computational model of our taxonomy (see Section~\ref{sec:3.2})
by relying on the mathematical tools employed. However, we argue that there still is a
need for this criterion in the developmental dimension of our taxonomy, as properly deciding which
action to take is a core aspect of developing sound and complete action knowledge:
\begin{itemize}
	\item \emph{classification} relates to agents which relate their perceptual input patterns
		to concrete categorical outputs. In this spirit, an agent identifies classes of actions
		which it implicitly relates to similar input patterns by defining a mapping from continuous
		to discrete spaces. Observe that classification transparently enables generalization (see Section~\ref{sec:3.1.2.4}; see rows 1--4 or 6--7 of Table~\ref{tab:class_cl}).
	\item \emph{regression} relates to agents which estimate the proper action to take given
		relations in its perceptual inputs. That is, given its stimuli an agent learns a regression
		function that maps from continuous to continuous spaces (see rows 9, 11 or 15 of Table~\ref{tab:class_cl}).
	\item \emph{inference} is a naturally inspired mechanism where an agent uses a set of acquired
		facts (existing knowledge) and hard-coded rules to infer new facts (novel knowledge), i.e., which
		action to take in a specific context. The rules may be represented as logical formulas,
		connections within graphs, or decision trees. Formally, this defines a mapping 
		from discrete to discrete spaces (see rows 14, 19 or 31 of Table~\ref{tab:class_cl}).
	\item \emph{optimization} is a purely mathematically-inspired mechanism to learn the best expected
		outcome given some input. Using it, an agent chooses an action that either maximizes a
		reward or minimizes a loss. Formally, this defines a mapping from either discrete or continuous
		to continuous spaces (see rows 5, 8 or 26 of Table~\ref{tab:class_cl}).
\end{itemize}

\paragraph{Learning}\label{sec:3.1.3.5}

Acquisition (see Section~\ref{sec:3.1.3.2}) of new information is an important capacity for
autonomous agents to avoid stagnation. However, acquisition is only part of the deal. An agent
also needs to be able to learn from this newly acquired knowledge in order to evolve. 
The means of learning are crucial for the development of both the agent and its internal
action model. Our taxonomy studies this criterion by two possible choices:
\begin{itemize}
	\item \emph{offline} learning characterizes agents that first acquire data (or are provisioned 
		with already-collected data) and subsequently employ this data
		for offline learning to acquire new knowledge. A drawback of this is that the agent
		may not be able to immediately react to changes in the environment or its embodiment, 
		or to validate the learning outcomes itself in the real world (see Section~\ref{sec:3.1.4.2}). 
		Yet, learning can be shaped more efficiently compared to online learning (see below; see rows 1--8 or 11--13 of Table~\ref{tab:class_cl}).
	\item \emph{online} learning poses novel challenges to an agent, i.e.,
		incomplete data and a large amount of noise and irrelevant data. That is, an agent, while
		exploring its environment to collect new data, is faced not only with the challenge to
		learn from this very data but also to filter out the relevant bits and pieces (\emph{c.f.} Section~\ref{sec:3.1.1.3}).
		Despite this disadvantage, online learning comes with the advantage of immediate adaptability
		to changes in both the environment and the embodiment (see rows 9--10 or 14--15 or 25 of Table~\ref{tab:class_cl}).
\end{itemize}

\paragraph{Acquisition}\label{sec:3.1.3.2}

To be able to learn something new an agents needs to be provided with information it is able to
process. Over the years, the robotics and machine learning community have drawn on various
formats of information provision for agents. Clearly, each of those come with their unique
advantages and disadvantages, which however are not the focus of this article. This criterion thus
does not study advantages or disadvantages of the means of information provision but instead how
the agent is provided with this novel information:
\begin{itemize}
	\item \emph{hard coded} implies that an agent generally does not acquire new knowledge but
		rather is provided with an initial set of, e.g., rules and facts about the world which
		allow it to shape its behavior. Clearly, such an agents stagnates until its knowledge
		base is manually extended (see rows 17, 48 or 52 of Table~\ref{tab:class_cl}).
	\item \emph{ground truth} implies that an agent acquires new knowledge by learning to
		relate specific input stimuli to actual outputs (e.g., motor commands) for achieving
		a desired effect. Agents thus are able to learn but only if provided with valid feedback
		on their choices. Observe that ground truth traditionally is a manually-specified feedback 
		signal that does not adapt to changes and may bias the learner (see rows 16 or 18--20 of Table~\ref{tab:class_cl}).
	\item \emph{demonstration} implies that an agent learns from another agent or human
		teacher by being instructed on how to perform specific actions. This kind of acquisition comes
		with the advantage that the agent can immediately relate what it is shown to itself resulting
		in more efficient learning (see rows 1, 3 or 5--8 of Table~\ref{tab:class_cl}).
	\item \emph{exploration} relates to agents that learn by exploring their environment by their
		own means, e.g., motor babbling. Being able to acquire new knowledge by exploring however
		requires the agent to be able to perceive and classify effects and changes in the world
		such that it can make sense of its actions (see rows 9, 14 or 33 of Table~\ref{tab:class_cl}).
	\item \emph{language} probably is the most difficult but also most advanced means of acquiring
		novel action knowledge. The format may have lots of different variations, from direct
		imperative instructions (which are arguably the easiest to understand) to scene
		explanations from which the agent is required to extract the relevant bits and pieces
		that describe the action it is observing and is supposed to acquire. Clearly, being able
		to learn actions by language is an advanced, high-level cognitive ability and thus hard 
		to achieve (see rows 22, 60 or 70 of Table~\ref{tab:class_cl}).
\end{itemize}
Observe that this criterion is again multi-choice, i.e., the means by which an agent acquires
new knowledge are not restricted to just one source (e.g., an agent may learn about new
actions by both being demonstrated what to do and at the same time being told what is actually
done; see rows 11, 58 or 86 of Table~\ref{tab:class_cl}).

\subsubsection{Effect}\label{sec:3.1.4} As already claimed by~\cite{Jeannerod2006},
in humans, actions are represented by their effects. Our taxonomy reflects this claim 
by containing a distinct dimension to study effect-related aspects of action models.
Clearly, our notion of effect does not immediately correspond to a ``mental'' representation of an action.
Nevertheless, it is an important aspect for studying the faithfulness of an action representation
and its underlying action model.

\paragraph{Discretization}\label{sec:3.1.4.1}

Effect discretization studies the granularity of effect predictions that an action model
supports. Effects may be either easily categorizable by clustering similar effects
or they may reside in a continuous spectrum. In our taxonomy, the discretization of effects
thence can fall into one of two categories:
\begin{itemize}
	\item \emph{categorical} effects generally relate to individual and different effects. Thence,
		effects under this category generally describe fixed amounts or clearly-distinguishable events
		as a result of performing an action. Observe that both numeral and symbolic effects are
		subsumed by this choice (see rows 2--4 or 7--8 of Table~\ref{tab:class_cl}).
	\item \emph{continuous} effects relate to fuzzy, boundless effects along a continuous
		dimension. Consequently, effects under this choice generally relate to real-valued action
		outcomes that are measurable along continuous spectra  (see rows 5, 9--10 or 12 of Table~\ref{tab:class_cl}).

\end{itemize}
Observe that this is a multi-choice criterion, i.e., an agent may as well consider both
categorical and continuous effects for establishing an action model (see rows 1, 5 or 37 of Table~\ref{tab:class_cl}).

\paragraph{Grounding}\label{sec:3.1.4.2}

Grounding of effects relates to the circumstance whether an action has or has not been executed in a 
real world environment by observing the intentional effects at the same time. Obviously, this criterion
is of utter importance as it expresses the maturity of an action model. If once executed in a real-world setting with the intended effects observed, the action is both feasible and
properly represented, whereas if not (i.e., only executed in simulation) one cannot guarantee that
an action is actually doable as intended. Thence, grounding binds intended effects to observable real-world events. 
Thus, grounding is either present or not (see rows 1--2, 5--7 or 9, and 3--4, 8 or 10--13, respectively, of Table~\ref{tab:class_cl}).

\paragraph{Associativity}\label{sec:3.1.4.3}

Associativity of effects relates to the capacity of both predicting the effects of an action
as well as predicting a necessary action to achieve a desired and intentional effect~\citep{Paulus2011}.
More precisely, this dimension does not directly investigate the mechanism for such capacities
but instead whether the action model possesses this capacity and further, the nature of this capacity.
Effect associativity can fall into one of two categories:
\begin{itemize}
	\item \emph{unidirectional} action-effect associativity categorizes an action model as only being
		able to infer the effects of executing a specific action. Consequently, an action representation
		lacks the capability of imagining which actions to execute to achieve a desired effect. On the
		contrary, given an action the model is readily capable of predicting the effects (see rows 2--13, 17 or 19--20 of Table~\ref{tab:class_cl}).
	\item \emph{bidirectional} action-effect associativity categorizes an action model as possessing
		the capacity to predict relevant actions given some desired effect. This is ultimately related
		to mirror neurons which upon observation of an action (that involves and object) immediately 
		activate neural populations relevant for motor control. This immediately allows for mental 
		simulation of actions. However, observe that imagining does not readily trigger a 
		representation (\citealt{Elsner2001,RizzolattiL2001,Rizzolatti2004}; see rows 1, 14--15 or 36 of Table~\ref{tab:class_cl}).
\end{itemize}

\paragraph{Effect Correspondence}\label{sec:3.1.4.4}

As argued by \cite{Newton2017}, usually we exercise an action to achieve a desired effect. 
Here we argue that one needs to distinguish between the actual frame of reference, or correspondence, of 
the effect. On the one hand, an effect may relate to changes in the environment, that is, 
displacing some object or opening a drawer. However, desired effects may also relate to changes in 
one's own bodily configuration, consequently treating the change in the environment as a
consequence of the bodily change (\emph{c.f.} \citealt{OShaughnessy1973}; Section~\ref{sec:2}). Thence, the 
latter does not exclude changes in the environment but rather treats them as an indirect 
effect of executing an action triggered by the bodily effect. This criteria allows for three choices, 
viz.\ \emph{environment} and \emph{body} or the combination of both (see rows 4 or 9, and 1--2 or 5--7, and 3 or 8, respectively, of Table~\ref{tab:class_cl}).

\subsection{Computational model criteria}\label{sec:3.2}

Computational model criteria serve to assess implementational aspects of an
action representation by how characteristics of the action model are realized. Thence,
the computational model discusses the mathematical and theoretical underpinnings
of action representations. 

\subsubsection{Formulation}\label{sec:3.2.1}

Here we consider whether a computational model is mathematically or bioglogically motivated. Clearly, 
there is a strong overlap between both categories, as, e.g., nature has inspired countless learning 
algorithms. Thus the question of where we draw the exact line between mathematical and biological 
motivation is valid. Our answer to this question is that a mathematically-formulated model solely draws on
mathematical tools without the claim of being biologically plausible, whereas a biologically-inspired, or
 biomimetic, model aims at grounding its workings in biological and neural processes:
\begin{itemize}
	\item \emph{mathematical} implies that a computational model is purely relying on existing
		mathematical tools with no claim to be biologically inspired (see rows 1--11 or 13-20 of Table~\ref{tab:class_cl}).
	\item \emph{biomimetic} implies that a computational model uses biology and cognition as
		a precursor for selecting proper mathematical tools. Such models thus are inspired from
		biology and neuroscience (see rows 12, 21 or 33 of Table~\ref{tab:class_cl}). 
\end{itemize}

\subsubsection{Implementation}\label{sec:3.2.2}

The implementational dimension of an underlying computational model of an action representation
studies relevant aspects of the programmatic implementation. This subsumes (i) the
concrete mathematical tools that are employed for learning and prediction, (ii) the
environmental features that are used by the model, and (iii) the kind of training that
is applied to the model, and thus entails a purely technical dimension.

\paragraph{Training}\label{sec:3.2.2.3}

The last dimension of the implementational aspects of the computational model of action 
representations studies the training used to train the predictive aspects of the developmental 
dimension of the action model (see~\ref{sec:3.1.3}). Our taxonomy supports the four most common 
types of training prevalent in robotics research:
\begin{itemize}
	\item \emph{unsupervised} learning relates to procedures where no -- direct or indirect -- feedback 
		signal is used to drive the learning process. Eventually this requires an agent
		to detect relevant statistical patterns in as well as the underlying structure of data without guidance.
		With respect to developmental robotics, this conceptually relates to the autonomous discovery
		of patterns or concepts from perceptual inputs in all available channels (exteroceptive and
		proprioceptive; see Section~\ref{sec:3.1.1.2}; see rows 3, 8 or 10 of Table~\ref{tab:class_cl}).
	\item \emph{supervised} learning refers to learning given concrete feedback signals. That is,
		each input datum comes with a label informing the agent  whether its prediction (or
		classification) was correct or not. Ultimately the agent learns to predict novel target
		values for previously-unseen inputs. Common drawbacks of this kind of training are under- or
		overfitting resulting from too little or biased training data  (see rows 1--2 or 4--7 of Table~\ref{tab:class_cl}).
	\item \emph{self-supervised} learning refers to agents capable of applying different views
		on data for learning patterns and concepts. Subsequently, one view, e.g., a specific
		sensor modality, is used to drive learning in another data view. For example, an agent
		may use clustering for learning low-level concepts in data (e.g., different obstacles).
		Subsequently, the cluster outputs are then used as target values for learning higher-level
		concepts using supervised learning (e.g., navigation). The term self-supervised refers to
		the supervision emerging from the learning agent instead of an external source  (see rows 15, 27 or 73 of Table~\ref{tab:class_cl}).
	\item \emph{semi-supervised} learning is a hybrid form of learning relying on techniques
		from supervised as well as unsupervised learning. It most naturally resembles human learning
		in that it is initially bootstrapped from supervised learning by a caregiver, followed by
		life-long, unsupervised learning by autonomous exploration (see rows 70, 87 or 111--112 of Table~\ref{tab:class_cl}).
\end{itemize}

\paragraph{Features}\label{sec:3.2.2.2}

To be able to make meaning of inputs in terms of computation, an action model requires 
extraction of features present in the inputs. Clearly, it may also directly rely on the 
inputs without any further processing. This criterion thus subsumes all kinds of representations 
from pixel intensities over salient points to features yielding from outputs of deep neural nets. 
Similar to the previous criterion this is also an open choice criterion, as again, the multitude 
of available and possible feature representations is too vast to be captured formally.

\paragraph{Method}\label{sec:3.2.2.1}

The method relates solely to the employed mathematical mechanisms that underpin the various 
perceptual, structural, developmental and effect-related aspects of the corresponding action model. 
It is an open choice criterion as providing choices for the multitude of mathematical tools that 
may be employed is too vast to be captured formally.

\subsubsection{Evaluation}\label{sec:3.2.3}

The last dimension of the computational model underpinning an action representation discusses
the means by which the action representation under study has been evaluated. The purpose of
this dimension is two-fold: first, it indicates whether a model is just a theoretical
musing or has practical relevance. Second, it indicates the maturity of a model. We thus
claim that this dimension is of substantial importance. The choices are:
\begin{itemize}
	\item \emph{benchmark} refers to action representations that compete with others in terms of being 
		evaluated on an unbiased, explicitly-devised data set. Doing so immediately allows comparing 
		representations with each other in terms of their representational and functional capacity.
		Benchmarks can fall into two categories distinguished by how the baseline is established. In one case, 
		the baseline is computed from a specially-devised training data set and compared against
		a test data set. In the other case, a baseline is established from the results of reference studies
		investigating the same hypothesis to be then compared against the own model using the same data
		as the reference studies (see rows 3--4 or 7--8 of Table~\ref{tab:class_cl}).
	\item \emph{real robot} implies that an action representation has been evaluated on a real,
		physical robot. Clearly, this kind of evaluation is the strongest one as it requires a
		model to be robust against real-world noise and to be able to deal with potentially-incomplete data  (see rows 1--2, 5--6 or 9--15 of Table~\ref{tab:class_cl}).
	\item \emph{simulation} categorizes models as having only been evaluated in a simulated
		environment. Clearly, such an evaluation is weaker as the inevitable physics approximations
		and imperfect noise models fail to catch a real-world environment. Thus, for action representations
		only evaluated in simulation one cannot assess much more than that they may be practically
		feasible but not whether they truly are or not (see rows 21--22, 26 or 39 of Table~\ref{tab:class_cl}).
	\item \emph{virtual reality} is a relatively recent type of evaluating, among others, action 
		representations~\citep{Zech2017}. It refers to a type of evaluation where a human agent 
		provides non-simulated interactions in an otherwise simulated environment with a simulated 
		agent (see rows 31 or 95 of Table~\ref{tab:class_cl}).

\end{itemize}
Observe that this is a multi-choice criterion, i.e., a computational model of an action
representation may well be evaluated in multiple settings, e.g., preliminary evaluation
in simulation with subsequent evaluation on a benchmark (see rows 91 or 110 of Table~\ref{tab:class_cl}).


\section{Selection and classification of papers}\label{sec:4}

Paramount to performing a systematic literature review together for categorizing papers 
is a carefully designed search and selection procedure. This section will thus introduce 
our systematic search and selection procedure for identifying papers relevant for classification. 
Additionally, we identify relevant threats of validity to our study. The resulting classification 
of action representations in robotics covered in the selected publications is then used in 
the next section to indicate the adequacy of the defined criteria and for further discussions (see Sections~\ref{sec:5} 
and~\ref{sec:6}).

\subsection{Selection of publications}

The selection of relevant, peer-reviewed, primary publications requires the definition of a search strategy
as well as paper selection criteria together with a selection procedure applied to the collected papers.

\subsubsection{Search strategy}

The initial search conducted to collect candidate papers was done automatically on December 1st, 2017 by
consulting the following digital libraries:

\begin{itemize}
  \item IEEE Digital Library (\url{http://ieeexplore.ieee.org/}),
  \item ScienceDirect (\url{http://www.sciencedirect.com/}),
  \item SpringerLink (\url{http://link.springer.com/}),
  \item SAGE (\url{http://journals.sagepub.com}), and
  \item Frontiers in Neurorobotics (\url{https://www.frontiersin.org/journals/neurorobotics}).
\end{itemize}
These libraries were chosen as they cover most of the relevant research on robotics. The search 
string was kept simple, i.e.,
\begin{center}
  \verb+action representation AND robot+
\end{center}
in order to keep the search general enough and to avoid missing any publications employing more 
precise terminology. Observe that the search was applied to all of the following search fields: 
(i) paper title, (ii) abstract, (iii) body, and (iv) keywords.  The search produced a set of 1575 
retrieved papers, thus a paper selection process was subsequently employed to further filter the results.

\subsubsection{Paper selection}

\begin{figure*}[htb]
    \centering
    \includegraphics[width=.6\textwidth]{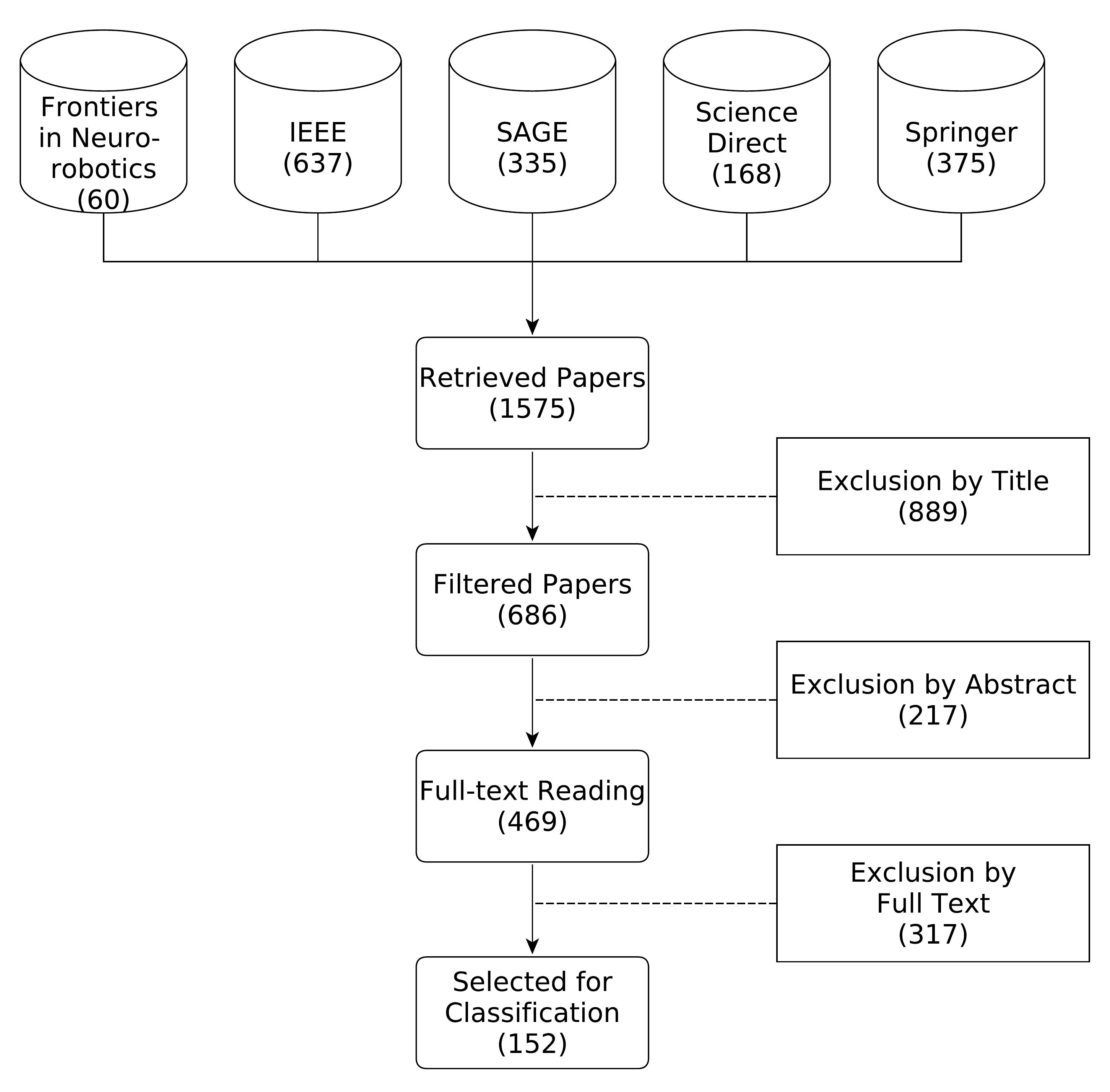}
    \caption{Selection of publications studied in this survey.}\label{fig:selection}
\end{figure*}

Figure~\ref{fig:selection} summarizes the paper selection process which comprised three phases.
In the first phase, papers were excluded based on their title: if the title did not indicate any
relevance to robotics and action representations, papers were discarded from the classification. 
This reduced the initial set of 1575 papers to 686 remaining papers. In the
second phase, papers were excluded based on their abstract, reducing the number of relevant papers to 
469. In the third and final phase, papers were rejected based on their content, reducing the 
set of relevant papers to 152. Thus, our classification, as discussed in Section~\ref{sec:6}, 
includes a total of 152 papers. Note that during the last iteration, a number of relevant 
papers were rejected on the basis that they either failed to introduce a novel representation or to 
sufficiently reevaluate an existing representation. Further, we deliberately excluded papers focusing 
solely on gesture recognition, as these generally are not considered mechanically-effective motions 
compared to actions~\citep{Hobaiter2017}.

\subsection{Paper Classification}

The 152 selected publications were categorized according to the classification criteria as 
defined and discussed in Section~\ref{sec:3} by four researchers. For this purpose, the remaining set 
of primary publications was randomly split into four sets of equal size for data extraction and classification. 
A classification spreadsheet was created for this purpose. Besides bibliographic information (title, authors, 
year, publisher) this sheet contains \emph{classification fields} for each of the defined criteria. To avoid 
misclassification, the scale and characteristics of each classification criterion were additionally implemented 
as a selection list for each criterion. As explained above, the list also contained the item ‘not specified’, 
to cater for situations where a specific criterion is not defined or could not be inferred from the contents of a 
paper. Problems encountered during the classification process were remarked upon in an additional comment field. 
The resulting classification of all publications was then reviewed independently by all four researchers. Finally, 
in multiple group sessions, all comments were discussed and resolved among all four researchers.

\subsection{Threats to validity}

Naturally there exist various issues that may influence the results of our study, e.g., the defined search string
as discussed previously. Threats to validity include multiple factors, most relevant to us (i) publication bias, 
(ii) identification and (iii) classification of publications, as well as the (iv) terminology employed.

\subsubsection{Publication bias}

This threat relates to the circumstance that only certain approaches, that is, those producing promising 
results or promoted by influential organizations are published~\citep{Kitchenham2004}. We regard this 
threat as moderate since the sources of publications were not restricted to a certain publisher, journal or 
conference. Therefore, we claim that our study sufficiently covers existing work in the field of action 
representations and robotics. However, to balance the trade-off between reviewing as much literature as 
possible while nevertheless accumulating reliable and relevant information, gray literature (technical 
reports, work in progress, unpublished or not peer-reviewed publications) was excluded~\citep{Kitchenham2004}. 
Further, the required number of pages was set to four to guarantee that publications contained enough information 
in order to categorize them appropriately.

\subsubsection{Threats to the identification of publications}

This threat is related to the circumstance that, during the search and selection of publications, relevant 
papers may have been missed. To address this, we employed a very general search string to avoid missing potentially 
relevant publications during the automated search. Yet, to additionally reduce the threat of missing important 
publications, we informally checked papers referenced by the selected papers. We did not become aware of any 
frequently cited papers that were missed.

Apart from that, we also want to point out that we deliberately excluded any papers discussing just plain 
reactive open- or closed-loop controllers, e.g., Dynamic movement Primitives (DMP) or Central Pattern Generators (CPG), 
as these, to the best of our knowledge, do not readily address the topic of action at a cognitive level but rather 
at the control level. Clearly, reactive control does not relate to the cognitive concept of an action being represented 
in terms of its effects and usually not readily coupled to some specific motor program. Additionally, we also excluded 
a large number of papers studying the application of reinforcement learning (RL). In general, RL assumes 
actions are already given (observe that we are interested in action representations and means of populating them by 
learning), and further, reinforcement learning also does not employ any notion of effect whatsoever.

\subsubsection{Threats to the classification of publications}

Given the rather large number of publications selected for classification according to a substantial number of 
defined criteria, the threat of misclassification needed to be addressed. Various measures were implemented in order 
to mitigate this threat. First of all, all criteria were precisely defined, as presented and discussed in Section~\ref{sec:3}, 
prior to the commencement of the paper selection and classification process. There was scope for the refinement of 
the concepts by the researchers during the process, but this was restricted mainly to descriptive adjustments.
Secondly, for each of the criteria we added a list of possible selections in the classification sheet to avoid 
misclassification. Third, the classification was conducted in parallel by four researchers who are experts in the 
field and who repeatedly cross-checked the classification independently. Finally, weekly meetings were held by 
the four researchers to discuss and resolve any comments that arose during independent classification.

\subsubsection{Terminology}

We are aware that the way we use specific terminology, e.g., \emph{action} and \emph{motion}, or, \emph{learning} and 
\emph{inference}, or \emph{understanding} may not be perfectly in line with their use in other areas of research. 
However, this survey has been written with a robotics research background, which is why we stick to the terminology 
as used in this field. Thus, given both this circumstance and the fact 
that the notion of an action representation, at least for now, is not that wide-spread in robotics we took the liberty 
to rigorously decide on our own when to use which term and whether some representation is an action representation or not.
However, readers from different fields should not face any problems in properly interpreting the content of this work, as 
the terminology as used in robotics research---to a high degree---has been coined by relevant concepts from psychology 
and neuroscience. On the other side, we hope that our work stimulates a discussion about the state of the art of 
action representations in robotics to advance this field and contributes to the establishment of a common 
and well-defined terminology.


\section{Results and Discussion}\label{sec:5}

This section comprises the main contribution of this article by presenting and discussing the classification 
of the selected papers (see Section~\ref{sec:4}). The complete classification of all 152 papers by 
the introduced taxonomy (see Section~\ref{sec:3}) is shown in the appendix of this article (see Tables~\ref{tab:class_cl} and \ref{tab:class_fm} 
in Appendix~\ref{sec:app_c}) and is also available online\footnote{\url{https://iis.uibk.ac.at/public/survey/ActionRepresentation/}}.

For each of the selected publications it was possible to categorize the presented action representation according to 
the criteria defined in Section~\ref{sec:3}. This indicates the pertinence of these criteria
for the classification of action representations in robotics, thence supplying a framework for understanding,
categorizing, assessing, and comparing action representations in robotics. Additionally, besides validating the 
criteria introduced in Section~\ref{sec:3}, our classification, having been conducted in a systematic and 
comprehensive manner, provides an aggregated view and investigation of current state of the art of action representations 
in robotics.

\begin{figure*}[htb]
  \centering
  \includegraphics[clip, trim=1.6cm 9.2cm 1.5cm 1.2cm, width=\textwidth]{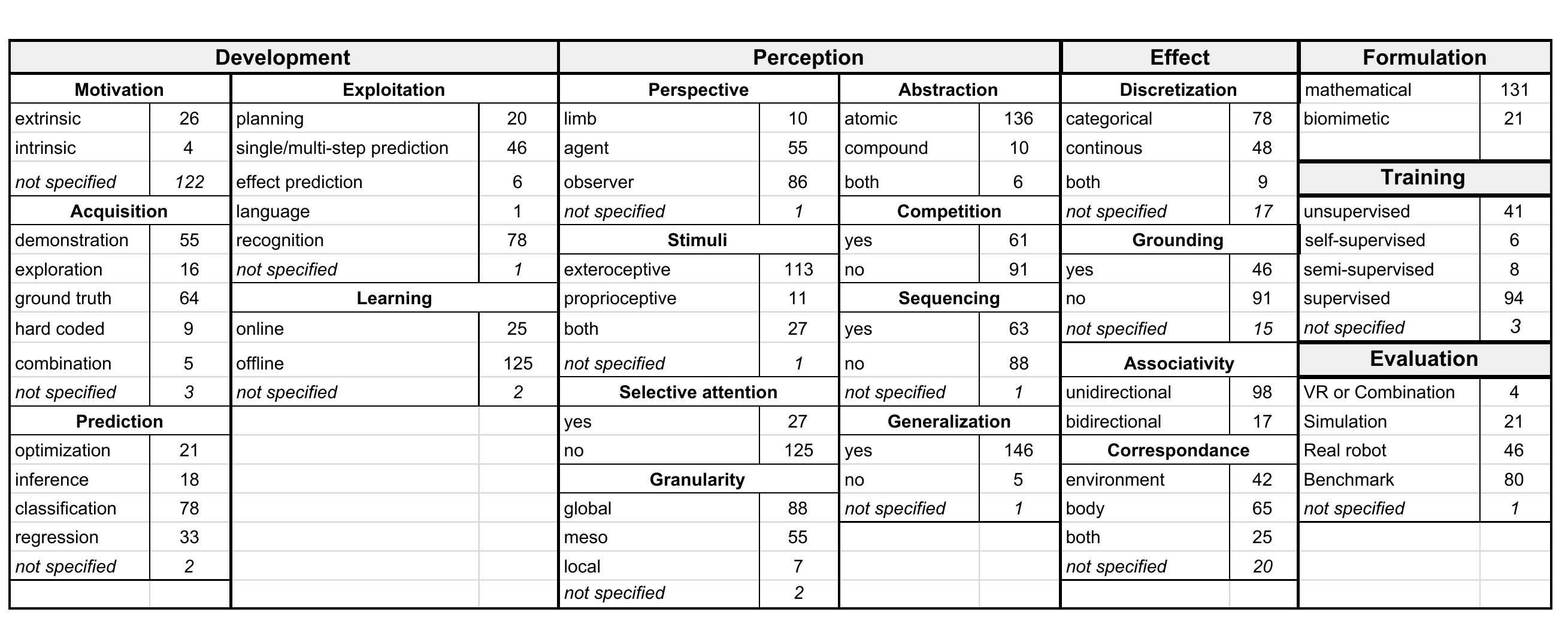}
  \caption{Numbers of papers falling into each category for all criteria.
  }
\label{fig:dist}
\end{figure*}

\begin{figure*}[htb]
  \centering
  \includegraphics[angle=90, clip, trim=1.6cm 6cm 1.5cm 1.2cm, width=0.73\textwidth]{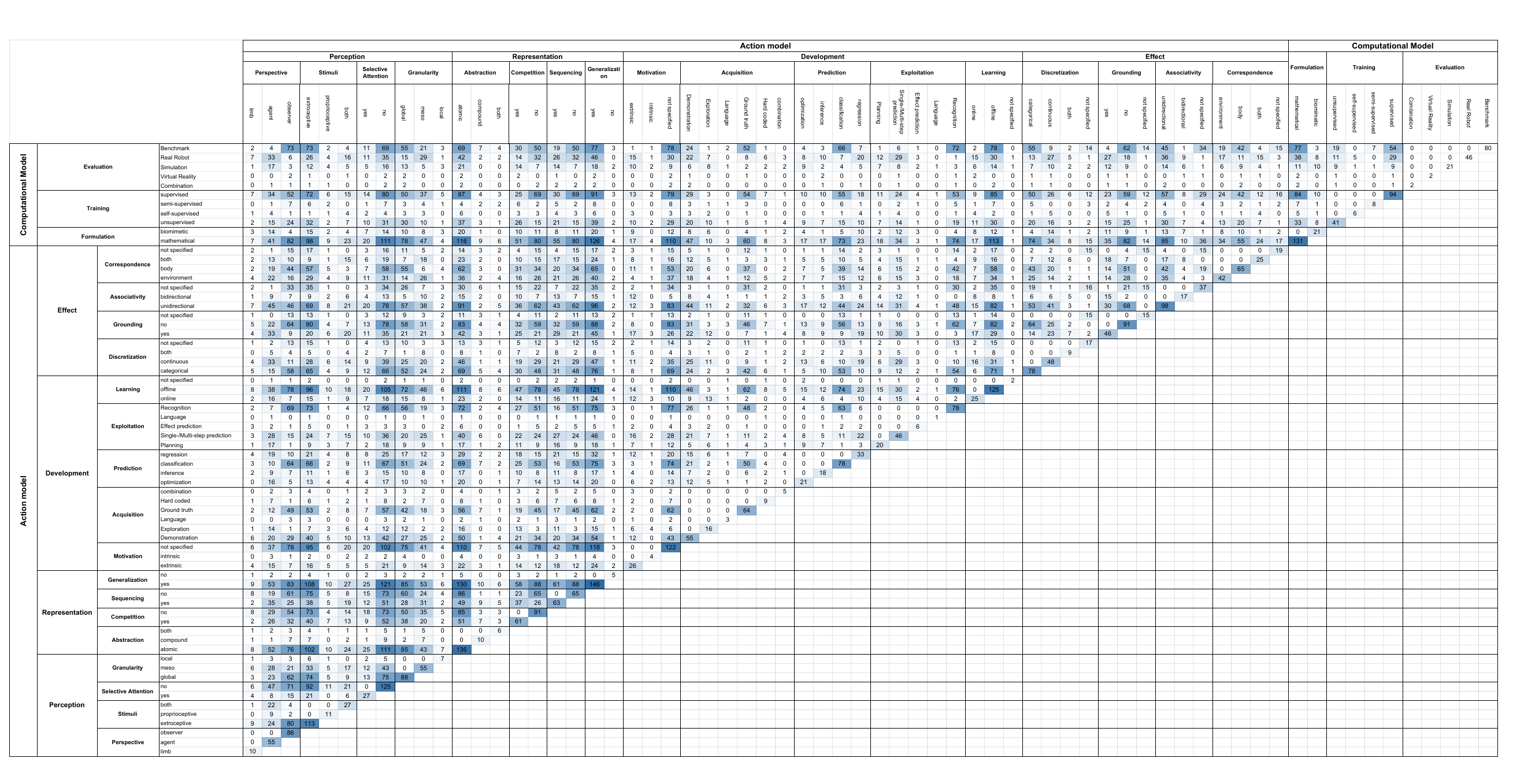}
  \caption{Co-ocurrence matrix of all criteria for all categorized papers (best viewed on a computer display; numbers missing for each criteria to sum to 152: not specified).\label{fig:cc}}
\end{figure*}

Figure~\ref{fig:cc} shows the summary statistics by a co-occurrence matrix of category values as defined in 
Section~\ref{sec:3} that arise in the analyzed papers, thus providing the foundation for subsequent discussions. Figure~\ref{fig:dist} gives the 
category distributions of the selected papers.

\subsection{Learning of Action Representations}\label{sec:5.1}

Learning, that is, the process of acquiring new or modifying existing knowledge, behaviors, skills, values, or 
preferences~\citep{Gross2015}, is one of the central aspects of action representations. Clearly, this usually 
requires proper motivation for learning to take place. Looking at Figure~\ref{fig:dist} shows that in great 
measure, the question of how to motivate (extrinsically or intrinsically) a learner is hardly addressed
(30 out of 152) and where it is, learning is chiefly extrinsically motivated (26 out of 30). Correlating this 
with the kind of training (see Figure~\ref{fig:cc}), we conjecture that this in general is because of the prevalence 
of supervised and offline learning (79 and 110 out of 152, respectively) which traditionally imposes the 
motivation of reducing some externally prescribed loss. In accordance to that, exploratory learning has also seen 
very little attention (16 of out 152). Clearly, such kind of learning would require switching to semi- or 
self-supervised online learning (1 and 4 of out 16 that do online learning). Furthermore, doing so would 
require a valid model of a robot's embodiment in order to learn what is possible given the available motor skills. 
In line with this, we also argue that manually-provided ground truth should be avoided as a means of a feedback 
signal for learning due to its static nature (69 out of 152). Using such manually-defined ground truth drastically 
impedes autonomous learning on a real robotic platform due to the dependence on teacher-dependent supervision (54 out of 69). 
Again, if learning is done on a real robotic platform we suggest the use of semi- or self-supervised online learning 
for immediate relation to the robot's embodiment.

The majority of the considered methods uses the observer perspective (86 out of 
152). Clearly, learning from such a perspective hinders the emergence of action representations for 
purposes other than plain recognition due to the yet-unsolved correspondence problem (\emph{c.f.} \cite{Zech2017}) and the 
consequent difficulty of relating observed actions to one's own embodiment. Admittedly, one can learn from observation but 
only in combination with subsequent exploration. Yet, we did not identify any such paper. On the bright side, however, 
there is still a substantial number of approaches that learn from the agent's perspective (55 out of 152), though only 
14 of those acquire new knowledge by exploration and 20 by demonstration. This readily corresponds to the prevalent use 
of only exteroceptive stimuli (113 out of 152). Observe that this again drastically foils relation to an agent's 
own embodiment. 

Noteworthy further drawbacks we currently see in learning action representations are (i) a lack of employing selective 
attention (27 of out 152), (ii) scarcity of language use (3 out of 152), (iii) negligence of learning with 
reference to an agent's limbs (10 out of 152), and (iv) only considering discrete instead of continuous (or both 
discrete and continuous) effects (78 out of 152). Obviously, selective attention allows the curse of dimensionality to be tackled by 
focusing on what is relevant. Further, learning with respect to the limbs eases re-execution of trained actions due to the simplified
planning problem, i.e., there is no need to do whole-body planning. Thirdly, using language enhances structuring and 
understanding of action knowledge thanks to the tight relation between language and action~\citep{Guerra-Filho2007}. Finally, 
enabling agents to reason about not only discrete but also continuous effects unlocks the ability to plan with respect to local 
changes in both the environment and the embodiment, and not only at a global environmental scale.

To conclude, in the area of learning action representations, the current multiple drawbacks stem in general from the 
prevalent combination of supervised, offline learning from an observer's perspective. We suggest that in the future, online learning 
in a semi- or self-supervised way from the agent's perspective merits more emphasis to resolve issues like the correspondence 
problem or proper motivation for learning.

\subsection{Maturity of Action Representations}\label{sec:5.2}

Two of the central criteria of our taxonomy directly relating to the maturity of an action representation are the means of 
exploitation and evaluation. Clearly, representations that allow only for recognition and that further are only evaluated on a 
benchmark lack maturity, missing empiricism yielding from real-world experiments on an actual robotic platform. In this respect, 
Figure~\ref{fig:dist} draws a rather disappointing picture in that more than half of the categorized papers have only been evaluated 
in terms of benchmarks (80 out of 152). Correlating this to the type of exploitation (see Figure~\ref{fig:cc}) we see that the 
bulk of these papers (72 out of 80) only do recognition. The main drawback coming along with such methods is the use of only 
exteroceptive stimuli and features which undermine construction of internal representations of one's own embodiment due to the missing 
relation between observation and embodiment. Yet, as neuroscience claims, such representations of one's embodiment are paramount for action 
recognition. Only with such models of the self at our disposal we are able to map observed actions onto our own 
embodiment for reexecution~\citep{SokolovGTP2009}. This mapping is crucial as it immediately solves the the 
correspondence problem. On the other side, the nastiness of the correspondence problem in combination with lacking 
representations of the self (and thereof emerging relations to an agent's embodiment) immediately explains why the works that address 
action recognition fail to close the gap towards re-execution of observed actions.

Another problem that emerges if looking closer at the plethora of papers doing action recognition is their stopping short 
of action sequencing (0 out of 72 address action sequencing). However, looking at Figure~\ref{fig:dist} immediately 
reveals that papers not focusing on recognition but rather on single- and multi-step prediction as well as planning are capable of 
sequencing actions (27 and 16 out of 63). Unfortunately however, these methods only allow sequencing single actions together 
but fail to represent resulting action sequences as compound actions. In contrast, those action representations that are able to handle 
compound actions (4 out of 152) do not address sequencing of such compound actions. 

The ability to handle action competition in the representation is another key aspect regarding the maturity of a model. Clearly, 
in every situation an agent is faced with multiple actions that yield similar or identical effects; thus it has to choose which action, 
among the feasible ones, to ultimately execute. In total, however, the number of approaches able to handle competition is less than half 
of all the papers we categorized (61 out of 152). Yet, using proper mathematical mechanics one actually can get competition for free, e.g., 
by employing neural networks or any other type of regressor/classifier that intrinsically handles competition at the decision level. 
However, we see a further potential reason for this general lack of handling competition, motivated by the circumstance that most 
works are only able to handle a couple of actions, possibly rendering competition useless for now. Yet, future work should put more 
emphasis on action competition, rendering agents more autonomous.

A last but very important indicator for the maturity of an action representation is the way it represents and handles effects. 
In general, the works we categorized focus on categorical effects (75 out of 152 papers) with only unidirectional associativity 
(95 out of 152). Correlating this to the category of exploitation, we again see that the majority of the representations that 
are only able to handle categorical effects are exploited for action recognition (54 out of 75). Clearly this is due to only 
recognizing classes of actions but not the continuous changes that the effects yield in both the agent's embodiment and its 
environment. However, this substantial lack of handling continuous effects has a further reason: a shortcoming in grounding effects 
in the real world (46 out of 152). Real-world physics in general are not discrete but continuous dynamical systems. 
Only by verifying estimations by real-world observations can we expect an agent to truly learn about the effects that it 
causes as well as its potential control over its environment. Finally, a last major drawback from our perspective is the prevalent 
unidirectional effect association (98 out of 152). This immediately yields scarcity of inverse models for inferring what to do to 
achieve a desired effect, consequently reducing the autonomy of the agent.

To sum up, we submit that the majority of existing action representations are not in a very mature state. This follows from
three major observations. First, evaluation mostly is not done on real robotic platforms. Secondly, researchers for now mainly 
focused on constructing representations only for recognition that neglect the self. And third, for most of the categorized works 
effects are not grounded in real-world physical environments. By putting more emphasis on these issues we claim that existing 
drawbacks, e.g., the shortcoming of proper inverse models, could readily be addressed.

\subsection{Formalizing Action Representations}\label{sec:5.3}

One of the central yet quite disappointing insights of our systematic search and classification is the realization 
that in robotics, usually, there is no widespread use of specifically devised data types (think about an abstract data 
type) for storing and managing action-specific knowledge. Clearly, such data types are however necessary as our earlier 
treatise in Section~\ref{sec:2} shows where in general one can see strong arguments in favor of internal representations of 
both actions and the self (\emph{c.f.} \cite{Mandik2005,Jeannerod2006,TunikFG2005,Naito2016}). Yet, except for the work 
of Beetz~\etal \citep{Tenorth_2012,Bartels_2013}, as well as Wörgötter~\etal \citep{Worgotter_2013,Aksoy_2013,
Vuga_2015,Aksoy_2016a} there has been little effort towards the design of appropriate data structures for storing, accessing, 
and transferring action knowledge. Quite the contrary, what is done in most categorized papers is to leverage existing vision-based 
feature extractors (e.g., CNNs) and descriptors, and to subsequently use a combination of those as input to some regressor/classifier. 
Obviously these vision-based features and descriptors in general do not express anything related to a specific action except for maybe 
what it ``looks like'', but doubtlessly no information regarding how to actually perform the action (\emph{c.f.} our earlier 
writing on closing the gap between recognition and re-execution in Section~\ref{sec:5.2}). Apart from that, in respect of Searle's famous definition of 
a computer being \emph{a device that manipulates formal symbols} \citep{Searle1997}, we conjecture that for artificial 
agents, valid representations of both actions and the self are inevitable. Formal symbols are representations. So, at the end of the day, 
an artificial agent needs internal representations to be able to compute.

Since Francis' influential article on the internal principle of control theory \citep{Francis1976} it is generally accepted 
that one of the central pillars of mammalian motor cognition strongly builds on inverse models for motor control \citep{Wolpert1998}. 
In the course of our survey we identified exactly one paper out of 152 (see Table~\ref{tab:class_cl} and \ref{tab:class_fm} in the Appendix) that makes use of explicit inverse models for single-/multi step prediction. Obviously this astonishing ignorance 
of inverse models only fortifies what we already argued earlier regarding the maturity of action representations. Yet, this lack 
of inverse models readily can be tackled by carefully revising existing representations and their 
mathematical underpinnings. We claim that doing so is paramount to verily advance the current state of the art in action representations in 
robotics. From a present-day perspective, in the long run this would also aid in effect modelling for action representations, as one 
readily obtains bidirectional effect associativity which currently is only addressed by a fraction of all categorized papers (17 out of 
152). We guess that the concurrent absence of inverse models as just discussed is further fostered by also not attributing 
neuroscientific results enough consideration in terms of building biomimetic models for action representations (21 out of of 152).

Another blind spot we revealed in the context of formalizing action representations is that, to a great extent, model formalizations 
are only done at the subsymbolic level. That is, looking at Tables~\ref{tab:class_cl} and~\ref{tab:class_fm} one sees a strong predominance of methods that 
purely operate at a subsymbolic level by means of the used features. Clearly, higher-level cognition requires symbolization of 
acquired knowledge for high-level abstract task planning. The results of our classification as shown in Figure~\ref{fig:dist} reinforce 
our observation in that only a small fraction of categorized action representations are exploited for high-level task planning (20 out 
of 152). We argue that action representations require proper symbolization for unlocking high-level abstract task planning.

Finally, a last point to discuss in the context of action representation formalizations is the scant use of optimization (21 out of 
152). We argue that optimization should be a first-class choice as ultimately one wants to optimize behavior by choosing the most 
fitting action. Correlating these papers to the kind of exploitation we at least see that 8 
out of those do single-/multi-step prediction, and 9 do planning, respectively, indicating that if optimization is used, then it is 
for optimizing behavior. Nevertheless, we argue that more emphasis should be put on optimization for action selection and behavior shaping. 
Observe that this however does not call for an increased use of RL at this point. RL in general is not about optimizing an action but 
rather the sequence of actions that is taken to fulfill a task. Optimization of the action itself should take place before policy optimization.

\subsection{Usability of Action Representations}\label{sec:5.4}

One of the paramount questions when talking about formal models in a general sense is their usability. The Oxford English dictionary 
defines usability as \emph{the degree to which something is able or fit to be used}. Now, this definition is very broad and does 
not really investigate what it means to be usable or how to actually measure whether something is usable. Let us therefore expand 
this definition by introducing three characteristics that we consider relevant for quantifying the usability of an action representation:
\begin{itemize}
	\item \emph{effectiveness}, i.e., the completeness and accuracy of a representation
	\item \emph{efficiency}, i.e., how long does a representation need to be learned and also how easily can it 
		be leveraged for executing a desired action
	\item \emph{robustness}, i.e., how well does the representation generalize, but also deal with incomplete/corrupt 
		data
\end{itemize}

Regarding effectiveness we clearly see a large shortcoming in currently-available action representations. Looking at 
Figure~\ref{fig:dist} (and as already mentioned) the bulk of existing methods solely do action recognition (78 out of 152). 
Despite being aware that recognition capabilities are crucial for action representations, we however claim that this is only 
the first step towards more powerful representations that also allow for motor imagery and actual execution of the abstracted 
action. Especially single- and multistep prediction is of high importance (46 out of 152) due to its immediate relation to 
deciding what to do next. Unfortunately however, this again boils down to closing the gap between recognition 
and execution (as already mentioned) as well as the correspondence problem for properly learning from demonstration. Further, 
this also comprises consideration of continuous effects for being able to come up with precise and accurate predictions 
regarding dynamic changes in the environment. 

Regarding efficiency we submit that current models are learnable with reasonable expense, at least in the event of supervised,
offline learning (85 out of 152). However, one has to keep in mind the general shortcoming of such models in that they 
generally only allow for action recognition (53 out of 85). Clearly, one has to keep in mind that in the case
of exploratory, self- or semi-supervised learning, learning a representation will take substantially longer. Unfortunately, as
our survey shows, exploratory learning has not been sufficiently addressed for learning action representations (16
out of 152). Observe that this lack of exploratory learning immediately relates to the maturity of a model by means of whether a representation is evaluated 
on a real robot or not. Clearly, learning and evaluating action representations on real robotic platforms strengthens the maturity 
of a representation. 

One of the hallmark features of the human mind is its robustness to noisy or corrupt sensory inputs. This 
capacity stems for one central feat of human development: lifelong learning in a noisy and dynamic environment. Hence, only by grounding observations 
in real-world experiences, our minds are able to develop robust motor control \citep{Harnad1990}. It is thence evident that for 
action representations in robotics we conjecture that such robustness yielding from grounding experiences in real-world 
observations is paramount. Besides, the capacity to generalize to new situations also plays a major role when it comes to 
robustness. Obviously, not being able to generalize to novel situations likely indicates a very weak model. Looking at 
Figure~\ref{fig:dist} we see that nearly all categorized methods generalize to novel situations (146 out of 152) indicating 
high robustness of most approaches. Yet, looking at how many of those ground effects shows quite a different picture. Not even a third of 
those (46 out of 146) actually ground effects by real-world experiences, hence now undermining the robustness of the 
remaining approaches. Correlating these numbers with the means of exploitation however immediately reveals that 73 of the 
models not grounding effects are exploited only for recognition (observe that the remaining three recognition 
models do ground effects). Undoubtedly, recognition is feasible without grounding effects. For the remaining 27 models we 
unfortunately either lack the relevant data, or, in the other case, these models mostly do single-/multistep prediction 
using models trained by video sequences. The above epitomizes again the prevalence of recognition models which just do not 
require effect grounding. In the remaining cases, we conjecture that this due to a neglect of selective attention 
(only 27 out of 152 do so). Naturally, selective attention allows the curse of dimensionality to be tackled by focusing only 
on the stimuli that are relevant, thereby catalyzing the grounding of effects. Figure~\ref{fig:cc} however reveals that 
only 11 out of those 27 models ground effects. We claim that future action representations need to capitalize on selective 
attention for facilitating effect grounding thus drastically improving robustness.

Compiling the above, usability is essential for action representations. Current issues as discussed however could be 
tackled by implementing and especially evaluating a representation directly on a real robotic platform. Such an approach 
immediately unlocks the grounding of effects and consequently strengthen the maturity of the evaluated representation. 
By additionally considering selective attention one readily ends up with a representation substantially more robust than 
most current approaches.

\subsection{A Few Last Words on Action and Activity Recognition Datasets}\label{sec:5.5}

Inspired by a recent survey of~\cite{Chaquet2013} we also investigated the use and wide-spread uptake of datasets as reported by the categorized
papers. Table~\ref{tab:dataset} shows the resulting distribution of datasets as reported by our classification. In total, 41 
different datasets have been used by various papers if evaluating an action representation using a benchmark (80 out of 152, see 
Figure~\ref{fig:dist}). Investigating the actual usage count of the various datasets, Table~\ref{tab:dataset} shows a similar 
preference pattern as Table~5 of \citeauthor{Chaquet2013}'s \citeyearpar{Chaquet2013} survey.  For example, KTH, Weizmann and IXMAS are all among the top five datasets 
used. If learning of action representations is possible from datasets for action recognition, evaluating the relevance of 
the representation for robotics should be similarily straightforward (\emph{c.f.} computer vision \citep{Wu2015,Russakovsky2015}). 
It is thus critical to define suitable, standardized datasets to learn action knowledge and corresponding benchmarking setups to 
properly evaluate the representation. This would greatly enhance quantitative comparison of different approaches, simply because 
the baseline is the same.

A more severe usage pattern is shown by Table~\ref{tab:dataset1} in that only a small fraction of papers evaluated on benchmark 
datasets used more than two datasets. Evaluating a model only on one or two datasets may drastically falsify results regarding 
generalization capabilities, simply because of focusing only on a small set of actions captured in just a couple 
of environments. Considering multiple datasets for evaluation---in line with the above---further allows for more insight 
into the behavior and capabilities of a model, and therefore for more robust models by virtue of better understanding.

We submit that applying more diversity in evaluating models on benchmarks, that is, using multiple and especially commonly 
used datasets, would greatly advance research on action representations in robotics. This advancement eventually capitalizes on 
deeper insight and understanding of how these various models actually achieve their desired outcome by meaningful quantitative 
comparisons.
\begin{table}[htb]
	\centering
	\begin{footnotesize}
		\begin{tabular}{|l|l|}\hline
		  	{\bf Dataset} & {\bf \# Usage} \\ \hline
			KTH~\citep{KTH} & 15\\ \hline
			Weizmann~\citep{Weizmann} & 13\\ \hline
			IXMAS~\citep{IXMAS} & 8\\ \hline
			MSR-Action-3D~\citep{MSR-Action-3D} & 7\\ \hline
			HMDB~\citep{HMDB} & 4\\ \hline
			3D Action Pairs~\citep{3D-ActionPairs} & 2\\ \hline
			50 Salads~\citep{50-Salads} & 2\\ \hline
			ADLs~\citep{ADLs} & 2\\ \hline
			CAD-60~\citep{CAD-60} & 2\\ \hline
			CMU-MoCap~\citep{CMU-MoCap} & 2\\ \hline
			Florence3D Actions~\citep{Florence3D-Action} & 2\\ \hline
			HDM05~\citep{HDM05} & 2\\ \hline
			Hollywood2~\citep{Hollywood2} & 2\\ \hline
			MoPrim~\citep{MoPrim} & 2\\ \hline
			MSR-II~\citep{MSR-II} & 2\\ \hline
			MSR Daily Activiy~\citep{MSR-Daily-Activity} & 2\\ \hline
			UTKinect-Action~\citep{UTKinect-Action} & 2\\ \hline
			YouTube~\citep{YouTube} & 2\\ \hline
			UCF-101~\citep{UCF-101} & 2\\ \hline
			UCF-Sports~\citep{UCF-Sports} & 2\\ \hline
			Berkeley-MHAD~\citep{Berkeley-MHAD} & 1 \\ \hline
			ChaLearn Gesture~\citep{ChaLearn-Gesture} & 1\\ \hline
			CHEMLAB corpus~\citep{CHEMLAB-corpus} & 1\\ \hline
			FBG~\citep{FBG} & 1 \\ \hline
			Fish-action~\citep{Fish-action} & 1 \\ \hline
			G3D~\citep{G3D} & 1 \\ \hline
			Human Grasp~\citep{Human-Grasp} & 1 \\ \hline
			JIGSAWS~\citep{JIGSAWS} & 1 \\ \hline
			ManiAc~\citep{ManiAc} & 1 \\ \hline
			MSRC-12~\citep{MSRC-12} & 1\\ \hline
			MuHAVi~\citep{MuHAVi} & 1\\ \hline
			Olympic-Sports~\citep{Olympic-Sports} & 1\\ \hline
			Ravel~\citep{Ravel} & 1\\ \hline
			RGBD-HUDAACT~\citep{RGBD-HUDAACT} & 1 \\ \hline
			Reading Act~\citep{Chen_2014} & 1 \\ \hline
			Robust~\citep{Robust} & 1\\ \hline
			Stanford-40 Actions~\citep{Stanford-40-Actions} & 1\\ \hline
			SYSU-3D-HOI~\citep{SYSU-3D-HOI} & 1\\\hline
			TACoS~\citep{TACoS} & 1\\ \hline
			UMD~\citep{UMD} & 1\\ \hline
			UT-Interaction~\citep{UT-Interaction} & 1 \\\hline
			YouTube~\citep{YouTube} & 1\\ \hline
		\end{tabular}
	\end{footnotesize}
	\caption{Datasets used for benchmarking in various categorized papers with 
		respective usage count.}\label{tab:dataset}
\end{table}

\begin{table}
	\centering
	\begin{footnotesize}
		\begin{tabular}{|l|l|}\hline
			{\bf \# Datasets} & {\bf \# Papers} \\ \hline
			4 & 5 \\ \hline
			3 & 7 \\ \hline
			2 & 13 \\ \hline
			1 & 31 \\ \hline
		\end{tabular}
	\end{footnotesize}
	\caption{Total number of datasets used by various categorized papers.}\label{tab:dataset1}
\end{table}


\section{Open research challenges}\label{sec:6}

Our classification and the resulting discussion from the previous section show that action representations in robotics 
have been intensively studied in recent years. However, our discussions from Sections~\ref{sec:5.1}--\ref{sec:5.5} also 
reveal that the current state of the art regarding action representations in robotics is still in an early stage and 
currently suffers from multiple issues. Below we provide an overview of the central research challenges as revealed
by the results of our analysis. We believe that addressing these is paramount to successfully  advance research 
on action representations in robotics.
\begin{itemize}
	
	\item \emph{Intensifying effect-centricity and effect grounding} Grounding of effects in real-world percepts is 
		one of the key challenges from our point of view. Clearly, due to the vast amount if information available 
		at each moment from both the self and the environment this is a hard challenge. Yet, doing so is critical 
		to improve the quality of a model. As mentioned below, selective attention is one of the keys in handling this 
		vast amount of data. Yet, we further claim that the capability of processing multi-modal percepts 
		also substantially catalyzes the grounding of effects.
	\item \emph{Coupled Forward and inverse models} One of the central advantages of biomimetic models, 
		especially in the field of motor control and thence action representations, is their postulation of the 
		need for inverse models. It is thence necessary to carefully reconsider current results in neuroscience 
		and motor cognition (\emph{c.f.}~Section~\ref{sec:2}) to tackle the prevalent lack of inverse models. 
		Doing so, among other benefits,	readily unlocks the capacity of bidirectional effect associativity as well as 
		performing motor imagery \citep{Jeannerod2006}.
	\item \emph{Exploiting language for action understanding} The compositional and semantically-rich nature of 
		language is a strong prior for action understanding. Language provides precise and unambiguous semantics when 
		it comes to describing actions. Therefore, we claim that besides grounding of effects in real-world observations, rooting 
		the meaning of an action in natural language further boosts both learning and properly understanding an action. In 
		the long run, this allows learning of more abstract, i.e., disembodied, and thence useful action representations.
	\item \emph{Intrinsically-motivated, exploratory, semi- and self-supervised learning} Importantly, humans learn by 
		observation and subsequent exploration and interaction with their environment. Following this central motive, 
		it is crucial to allow computational agents to learn relevant concepts with minimal prior information. This allows for 
		progressive learning of representations of the external world as well as of the self. Clearly, this 
		requires an agent to be accordingly motivated as well as the capacity of self-supervising its learning 
		efforts. This ultimately culminates in using already-learned concepts, to both drive and supervise the learning autonomously. 
		We claim that learning in such a way yields stronger autonomy compared to classic supervised learning and thence merits more attention.
	\item \emph{Selective attention} Again, we argue similarly to \cite{Zech2017} that selective attention is an 
		important aspect for focused perception by blocking out clutter and noise. Contrary to our reasoning in the 
		case of affordance however, here we claim that selective attention should be ascribed a central role as a 
		precursor for grounding effects by successfully tackling the curse of dimensionality by only considering 
		those stimuli which are relevant for grounding the observed effects, thus drastically boosting the 
		robustness of different representations. Observe the immediate complementarity to the above challenge regarding 
		effect centricity and grounding of effects.
	\item \emph{Solving the correspondence problem} Similarly to \cite{Zech2017} we claim here that 
		it is of utmost importance to solve the correspondence problem in robotics, i.e., mapping of observed motions.
		This would address current drawbacks in both learning from demonstration and in understanding actions from 
		an observer’s point of view. Especially in the event of action representations this would allow closing the 
		gap from recognition to re-execution. Observe that this also requires intensified research towards
		constructing internal models of the agent's self.
	\item \emph{Sequence-based modeling} The capability of composing compound actions, e.g., pick-and-place, out of 
		more granular, atomic actions is a central capacity of mammalian motor control. Our minds do not store 
		complete motor programs for each and every action but rather dynamically synthesize them out of more 
		general building blocks for seamless action execution (\emph{c.f.}~Section~\ref{sec:2}). Clearly, such a capacity is 
		also paramount for action representations in robotics especially with regards to generalizability but 
		also scalability at a computational level.
\end{itemize}

Observe that there exists a substantial intersection of the above challenges with those identified by \cite{Zech2017} in the 
case of affordance research in robotics. This however is not surprising given the strong relation between actions and affordances,
the latter being a key driver in action selection. This intersection clearly resembles the strong interrelation of these two 
complementary fields of research and thus motivates joint research efforts.


\section{Conclusion}\label{sec:7}

Action representations are a key ingredient of autonomy in robots. In this article we thus made three major contributions 
relevant for this field of research. After a thorough survey of the meaning of action as well as contemporary definitions and 
opinions from various associated scientific disciplines we ended with a seminal definition of action relevant to robotics
(\emph{c.f.} Section~\ref{sec:2}). This treatise thence paved the way for the first major contribution of our article,
a taxonomy of action representations in robotics (\emph{c.f.} Section~\ref{sec:3}). This allowed us to conduct our second 
major contribution, a systematic review of existing work on action representations in robotics. Identified publications 
subsequently were categorized using our taxonomy, yielding the results for our third contribution in the 
form of an in-depth discussion of existing research on action representations in robotics (\emph{c.f.} Section~\ref{sec:5}). 
This discussion finally culminated in the identification of key research challenges we deem fundamental for advancing research 
on action representations in robotics ({\emph{c.f.} Section~\ref{sec:6}).

Summarizing our work we report that for now one of the central drawbacks in action research in robotics is the crucial lack 
of a common notion of both action and action representation in robotics. However, this shall not raise the impression that current 
state of the art work is useless. On the contrary, existing results act both as a foundation and guidance towards how to advance 
action research in robotics. Accordingly, in Section~\ref{sec:6} we identified future courses of actions for action research 
in robotics. We believe that intensifying research in these fields prolifically unlocks novel motor-cognitive capabilities in 
autonomous agents towards both more autonomy and dexterity.


\appendix

\section{Abbreviations for Classification}\label{sec:app_a}
\begin{flushleft}
Tables~\ref{tab:abbrev_am} and~\ref{tab:abbrev_cm} show the various abbreviations as used in the classification 
depicted in Tables~\ref{tab:class_cl} and \ref{tab:class_fm}.
\end{flushleft}

\vspace{0.5cm}
\bottomcaption{Abbreviations for action model.}\label{tab:abbrev_am}
\begin{footnotesize}
\centering
	\begin{xtabular}{ll} \specialrule{0.8pt}{1pt}{1pt} 
		{\bf Abbreviation} & {\bf Definition} \\ \hline
		b  &  both \\ 
		n  &  No \\ 
		ns &   not specified \\
		y  &  Yes \\ \hline
		{\bf Per}  & {\bf Perspective} \\
	        li  &  limb   \\
		ag  &  agent \\
		ob  &  observer \\ \hline
		{\bf St} &   {\bf Stimuli} \\
		e  &  exteroceptive \\
		p  &  proprioceptive \\\hline
		{\bf SA}  &  {\bf Selective Attention} \\ \hline
		{\bf Grn} & {\bf Granularity} \\
		lo  &  Local \\
		me  &  Meso \\
		gl  &  Global \\ \hline
		{\bf Abs} &  {\bf Abstraction}\\
		a  &  atomic \\
		c  &  compound \\ \hline
		{\bf Com} & {\bf Competition}\\ \hline
		{\bf Seq}  &  {\bf Sequencing}\\ \hline
		{\bf Mot}  &  {\bf Motivation} \\
		in  &  intrinsic \\
		ex  &  extrinsic \\ \hline
		{\bf Acq} &  {\bf Acquisition} \\
		com  &  combination \\
		d  &  demonstration \\
		exp  &  exploration \\
		gt &   ground truth \\
		hc &   hard coded \\
		l  &  language \\ \hline
		{\bf Pred} & {\bf Prediction}\\
		cla  &  classification \\
		inf  &  inference \\
		opt  &  optimization \\
		reg  &  regression \\ \hline
		{\bf Exp} & {\bf Exploitation}\\
		ep  &  effect prediction \\
		l   &  language \\
		p   &  planning \\
		r   &  recognition \\
		sa  &  self-assessment\\
		sp  &  single-/multi-step prediction \\ \hline
		{\bf Lrn} &   {\bf Learning} \\
		off  &  offline \\
		on   & online \\ \hline
		{\bf Disc} & {\bf Discretization}\\
		ca   & categorical \\
		co  &  continuous \\
		{\bf Gnd} &  {\bf Grounding}\\ \hline
		{\bf Asso} & {\bf Associativity}\\
		ud  &  unidirectional \\
		bd  &  bidirectional \\ \hline
		{\bf Corr} &  {\bf Effect Correspondence}\\
		by &  body \\ 
		env  &  environment \\ \hline
	\end{xtabular}
\end{footnotesize}
\vspace{0.5cm}
\bottomcaption{Abbreviations for computational model.}\label{tab:abbrev_cm}
\begin{footnotesize}
	\centering
	\begin{xtabular}{ll}\specialrule{0.8pt}{1pt}{1pt}
		{\bf Abbreviation} & {\bf Definition} \\ \hline
		{\bf Form} &  {\bf Formulation}\\
		MAT  &  mathematical \\
		BIO  &  biomimetic \\ \hline
		{\bf Train} &  {\bf Training} \\
		S  &  supervised \\
		SELF &   self-supervised \\
		SEMI &   semi-supervised \\
		U  &  unsupervised \\ \hline
		{\bf Eval} & {\bf Evaluation} \\
		BM  &  benchmark \\
		RR &   real robot \\
		SIM  &  simulation \\
		VR &   virtual reality \\
		C  &  combination \\\hline
	\end{xtabular}
\end{footnotesize}

\section{Abbreviations for Methods and Features}\label{sec:app_b}
\begin{flushleft}
Table~\ref{tab:abbrev_mf} lists the definitions of abbreviations denoting the various features and methods 
as reported by the papers categorized in Tables~\ref{tab:class_cl} and \ref{tab:class_fm}.
\end{flushleft}
\vspace{0.5cm}
\bottomcaption{Abbreviations for methods an features}\label{tab:abbrev_mf}
\tablefirsthead{\toprule \bf{Abbreviation} & \multicolumn{1}{l}{\bf{Definition}} \\ \midrule}
\begin{footnotesize}
	\centering
	\begin{xtabular}{ll}
	       	*MB& Model-Based\\
        	*MF& Model-Free\\
        	AE& Autoencoder\\
        	ANN & Artificial Neural Network\\
        	ASOM& Associative Som\\
        	BN & Bayesian Network\\
        	BOF & Bag-Of-Features\\
        	BOO & Bag-Of-Objects \\
        	BOW& Bag-Of-Words\\
        	BP & Belief Propagation\\
        	CMAC& Cerebellar Model Articulation Controller\\
        	CNN& Convolutional Neural Network\\
        	CRF& Conditional Random Field\\
        	CS& Conceptual Spaces\\
        	CTRNN& Continuous Time RNN\\
        	DAG-RNN& Directed Acyclic Graph RNN\\
        	DBN & Dynamic Bayesian Networks\\
        	DCNN& Deep CNN\\
        	DMP & Dynamic Movement Primitive\\
        	DNN& Deep Neural Network\\
        	DP & Dynamic Programming\\
        	DS& Dynamical System\\
        	DTW & Dynamic Time Warping\\
        	ECV & Early Cognitive Vision\\
        	EDM & Euclidean Distance Matrix \\
        	EKF& Extended Kalman Filter\\
        	ELM& Extreme Learning Machine \\
        	EM& Expectation-Maximization\\
        	EMG & Electromyography \\
        	FFT& Fast-Fourier Transform\\
        	FREAK & Fast Retina Keypoint \\
        	FSM& Finite-state Machine\\
        	FSTM & Feasible Situation Transition Manifold \\
        	GLOH & Gradient Location and Orientation Histogram \\
        	GMM& Gaussian Mixture Model\\
        	GMR& Gaussian Mixture Regression \\
        	GP& Gaussian Process\\
        	GPR &Gaussian Process Regression \\
        	GWR& Growing When Required Network \\
        	HHMM & Hierarchical Hidden Markov Model\\
        	HMM & Hidden Markov Model\\
        	HOG & History Of Gradients\\
        	HOS & Histogram Of Silhouette\\
        	HPNNA& Hierarchical Programmable NN Architecture\\
        	ICA & Independent Component Analysis\\
        	IMU & Internal Measurement Unit \\
        	KDE & Kernel Density Estimation\\
        	KD-Tree & k-Dimensional Tree\\
        	k-NN & k-Nearest Neighbor\\
        	LCSS& Longest Common Subsequence\\
        	LD & Levenshtein Distance \\
        	LSM& Liquid State Machine\\
        	LSTM& Long-Short Term Memory\\
        	LVQ & Learning Vector Quantization\\
        	MCSVM& Multiclass SVM\\
		MDN & Mixture Density Network \\
        	MDP & Markov Decision Process\\
        	MKL & Multiple Kernel Learning\\
        	MMI & Maximization of Mutual Information \\
        	MMM& Master Motor Map\\
        	MMR& Maximum Margin Regression\\
        	MNN& Modular Neural Networks\\
        	MoFREAK & Motion-Based FREAK \\
        	MSER & Maximally Stable Extremal Regions \\
        	MTRNN & Multiple Timescales RNN\\
        	NBNN& Naive Bayes Nearest Neighbor\\
        	NF & Neural Field \\
        	NGLD & Normalized Google-Like Distance\\
        	NLP& Natural Language Processing\\
        	NMF& Negative Matrix Factorization\\
        	NNC & Nearest Neighbour Classifier\\
        	NNMF& Non-NMF\\
        	NN & Neural Network \\
        	PCA& Principal Component Analysis\\
        	PCA-STOP & PCA Space-Time Occupancy Patterns \\
        	PDI & Positional Distribution Information\\
        	PHMM& Parametric HMM\\
        	PLSA & Probabilistic Latent Semantic Analysis\\
        	PMP& Passive Motion Paradigm\\
        	PMT & Projected Motion Template\\
        	PP& Purr-Puss\\
        	PSVM& Probabilistic SVM\\
        	PVS& Predicate Vector Sequence\\
        	QTC& Qualitative Trajectory Calculus\\
        	RBF& Radial Basis Function\\
        	RF& Random Forest\\
        	RGB-D & Red-Green-Blue-Depth \\
        	RGB & Red-Green-Blue \\
        	RL & Reinforcement Learning\\
        	RNNPB & RNN with Parametric Bias \\
        	RNN& Recurrent Neural Network\\
        	SCFG& Stochastic Context Free Grammar\\
        	SEC& Semantic Event Chain\\
        	SFA& Slow Feature Analysis\\
        	SIFT & Scale-Invariant Feature Transform\\
        	SOM & Self Organizing Map\\
        	SPHOF& Spatial Pyramid Histogram of Optical Flow\\
        	SSM & Self-Similarity Matrix\\
        	SSP & Space Salient Pairwise Feature \\
        	STDP& Spike-Timing Dependent Plasticity\\
        	STIP & Spatio-temporal interest points \\
        	STV & Spatio-Temporal Volumes\\
        	SVM& Support Vector Machine \\
        	SVR & Support Vector Regression\\
        	TSP & Time Salient Pairwise Feature \\
        	VAE & Variational AE \\
        	VMT & Volume Motion Template\\
        	WSM& Word Space Model\\
	\end{xtabular}
\end{footnotesize}

\section{Classification of selected publications}\label{sec:app_c}
\begin{flushleft}
Tables~\ref{tab:class_cl} and \ref{tab:class_fm} show the full classification of all selected publications. 
These results are also available online at \url{https://iis.uibk.ac.at/public/survey/ActionRepresentation/}.
\end{flushleft}
\include{classification-table}


\section*{Acknowledgments}

The research leading to these results has received funding from the European
Union’s Horizon 2020 research and innovation programme under grant agreement no.
731761, IMAGINE\@.


\bibliographystyle{SageH}
\bibliography{references}

\vspace*{-0.35cm}
\begin{biogs}
  \hspace*{0.2cm}
  \begin{minipage}{0.3\columnwidth}
    \includegraphics[width=0.9\columnwidth]{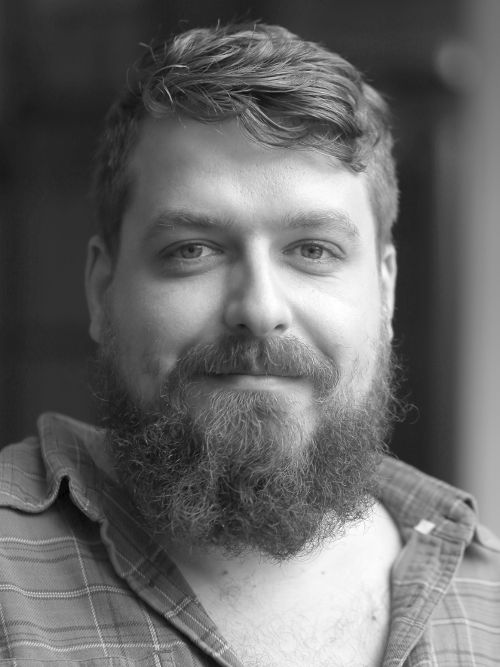}
  \end{minipage}
  \begin{minipage}{0.7\columnwidth}
    {\bf Philipp Zech} is a postdoctoral researcher at the University of
	Innsbruck. He received his Ph.D. degree in Computer Science
	from the University of Innsbruck in 2014. He is interested in
	developmental and cognitive robotics, affordance learning and
	intelligent and adaptive manipulation. He has already published
	more than 30 papers in international journals and conferences,
	two of which have received best-paper awards. He recently joined the
	editorial board of Adaptive Behavior as an associate editor.
  \end{minipage} \hfill
  \medskip

  \begin{minipage}{0.3\columnwidth}
    \includegraphics[width=0.9\textwidth]{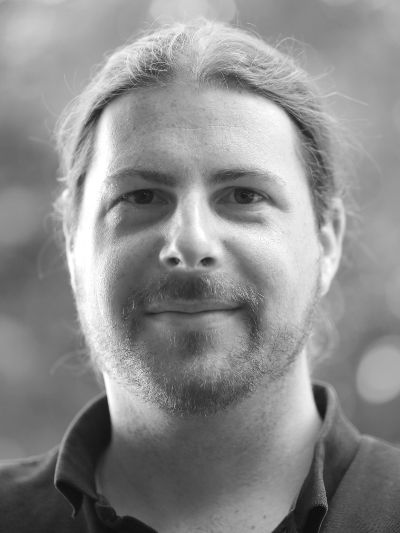}
  \end{minipage}
  \begin{minipage}{0.7\columnwidth}
    {\bf Erwan Renaudo} is a postdoctoral researcher at the University 
    of Innsbruck. He completed his PhD degree in 2016 at Pierre and 
    Marie Curie University where he worked on bio-inspired robotic 
    architectures with ensemble reinforcement learning. His research 
    interests focus on learning methods for autonomous robots. 
    He is particularly interested in autonomous behavior generation, 
    from action learning to coordination of habitual and goal-directed 
    behaviors.
  \end{minipage} \hfill
  \medskip

  \begin{minipage}{0.3\columnwidth}
    \includegraphics[width=0.9\columnwidth]{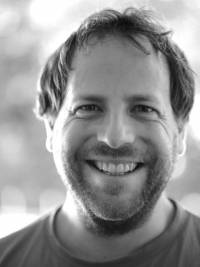}
  \end{minipage}
  \begin{minipage}{0.7\columnwidth}
    {\bf Simon Haller} is with the Department of Computer Science, University of
    Innsbruck, from where he received his B.Sc.~in 2006. In 2012 he was
    granted the professional title Ing.~by the Federal Ministry of
    Economics, Austria. He is a researcher and a scientific systems engineer.
    His work focuses on problem solving in the field of robotic hard- and
    software.
  \end{minipage} \hfill
  \medskip

  \begin{minipage}{0.3\columnwidth}
    \includegraphics[width=0.9\textwidth]{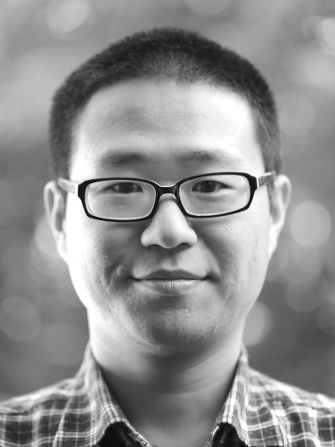}
  \end{minipage}
  \begin{minipage}{0.7\columnwidth}
    {\bf Xiang Zhang} Xiang Zhang is currently a PhD student at University of
     Innsbruck. He received his M.Sc. degree in System and Control from  Delft
      University of Technology in 2016. He is interested in robot learning from 
      demonstrations and intelligent and adaptive manipulation. 
  \end{minipage} \hfill
  \medskip

  \begin{minipage}{0.3\columnwidth}
    \includegraphics[width=0.9\columnwidth]{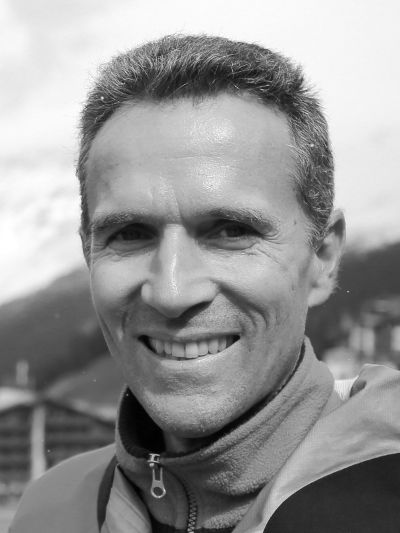}
  \end{minipage}
  \begin{minipage}{0.7\columnwidth}
	{\bf Justus Piater} Piater is a professor of computer science at the University 
	of Innsbruck, Austria, where he leads the Intelligent and Interactive Systems 
	group. He holds a M.Sc. degree from the University of Magdeburg, Germany, and M.Sc. 
	and Ph.D. degrees from the University of Massachusetts Amherst, USA, all in computer 
	science. Before joining the University of Innsbruck in 2010, he was a visiting researcher 
	at the Max Planck Institute for Biological Cybernetics in T\"ubingen, Germany, a professor 
	of computer science at the University of Liège, Belgium, and a Marie-Curie research 
	fellow at GRAVIR-IMAG, INRIA Rhône-Alpes, France. His research interests focus on 
	visual perception, learning and inference in sensorimotor systems. He has published 
	more than 170 papers in international journals and conferences, several of which 
	have received best-paper awards, and currently serves as Associate Editor of the 
	IEEE Transactions on Robotics.
  \end{minipage} \hfill
\end{biogs}

\end{document}